\newcommand{\ignoreForEBISS}[1]{#1}
\renewcommand{\GWord}[1]{}
\renewcommand{\ignoreForEBISS}[1]{}

\def \weights {\omega}     

\def \idxx   {i}           
\def \iidxx  {j}           
\def \Idxx   {I}           

\def \idxu   {\alpha}      
\def \iidxu  {\beta}       
\def \Idxu   {\Idxx}       

\def \IdxN   {M}           

\def \idxe   {l}           
\def \IdxE   {L}           

\def \numSubs{C}           

\def \eVecHLComp {\hat{w}}
\def \eVecHL {\vec{\eVecHLComp}}
\def \eValHL {\lambda}
\def \eVecLD {\vec{w}}
\def \eEleLD {w}
\def \eValLD {\eValHL} 
\def \eVecL  {\vec{u}}
\def \eEleL  {u}
\def \eValL  {\gamma}

\def \inDim  {N}           

\def \inVec  {\vec{x}}
\def \inMat  {\mat{X}}

\def \funcVec{\vec{f}}
\def \funcMat{\mat{F}}
\def \funcWVec{\vec{z}}

\def \funcDim{P}

\def \matUnderlineD{{\mat{D}^{-1/2}}}
\def \matUnderlineD{\mat{\underline{D}}}

\newcommand{\myProperty}[1]{$\langle$#1$\rangle$}
\newcommand{\mathProperty}[1]{\langle#1\rangle}

\begin{document}

\LNMaketitleAuthors{Laplacian Matrix for \\ Dimensionality Reduction and Clustering}{2017--2019}{Laurenz Wiskott \& Fabian Schönfeld \\ Institut f\"ur
Neuroinformatik \\ Ruhr-Universit\"at Bochum, Germany, EU}{
  %
18 September 2019
}

\tableofcontents


  \vspace{2ex}

  \paragraph{Requirements:} I assume the student can already ...
  \begin{itemize}
  \item[...] apply basic concepts from linear algebra, such as \emph{vector},
    \emph{matrix}, \emph{matrix product}, \emph{inverse matrix}.
  \item[...] solve an \emph{ordinary eigenvalue equation} in linear algebra and explain
    intuitively what \emph{eigenvalues} and \emph{eigenvectors} are.
  \item[...] relate the \emph{eigenvalues} and \emph{eigenvectors} of a symmetric matrix
    to the solutions of the minimization/maximization problem of the corresponding
    \emph{quadratic form}.
  \item[...] interpret a \emph{system of linear differential equations with constant
      coefficients}.
  \end{itemize}
    
  \paragraph{Learning objectives:} The learning objective of this unit is that the
  student can ...
  \begin{itemize}
  \item[...] define basic notions of graph theory, namely \emph{graph}, \emph{node},
    \emph{edge}, and \emph{simple graph} (Sec.~\ref{sec:SimpleGraph}).
  \item[...] explain matrix representations of graphs, namely \emph{adjacency
      matrix}, \emph{degree matrix}, and \emph{Laplacian matrix}
    (Sec.~\ref{sec:MatrixRepresentation}).
  \item[...] reproduce and interpret the generalized eigenvalue
    equation~(\ref{eq:GeneralizedEVE}) of the Laplacian matrix and weighted degree matrix
    and describe how it relates to the optimization problem of Laplacian eigenmaps and
    spectral clustering (Eqs.~\ref{eq:minuLu}--\ref{eq:orderDecorrD}).
  \item[...] summarize and motivate mathematical properties
    \myProperty{\ref{item:DOrthogonal},\ref{item:oneVector},\ref{item:indicatorVectors},\ref{item:weightedZeroMean}}
    (Sec.~\ref{sec:MathProp}) of the eigenvalues and eigenvectors of the generalized
    eigenvalue equation~(\ref{eq:GeneralizedEVE}).
  \item[...] discuss the role of the normalization constraint~(\ref{eq:normalizedD})
    vs.~(\ref{eq:normalizedI}) (Sec.~\ref{sec:RoleOfNormalization}).
  \item[...] explain how a similarity graph can be generated from a set of data
    points (Sec.~\ref{section:simGraphs}).
  \item[...] explain how the Laplacian eigenmaps (LEM) algorithm (Sec.~\ref{sec:LEM})
    and spectral clustering (Sec.~\ref{sec:SpectralClustering}) work.
  \item[...] name a limitation of LEM and sketch how locality preserving projections
    (LPP) overcome it (Sec.~\ref{sec:LPP}).
  \end{itemize}
  \ignoreForEBISS{Please also work on the exercises and read the solutions to deepen
    these concepts.}
  

\LNLecture{1}{2}{}


\section{Introduction}

\markText{Many problems in machine learning can be expressed by means of a graph with
  nodes representing training samples and edges representing the relationship between
  samples} in terms of similarity, temporal proximity, or label information.
\markText{Graphs can} in turn \markText{be represented by matrices.}  A special example
is \markText{the Laplacian matrix}, which allows us to assign each node a value that
varies only little between strongly connected nodes and more between distant nodes.  Such
an assignment \markText{can be used to} extract a useful feature representation,
\markText{find a good embedding\footnotemark\footnotetext{A remark on terminology: We
    use \emph{assign/assignment} for giving data samples an associated value.  These
    values implicitly define a \emph{mapping} from (possibly high-dimensional or
    non-vectorial) data samples to points in a low-dimensional space, the \emph{mapped
      space}.  In LPP the mapping is defined more explicitly by a linear function.  The
    collection of points in mapped space form an \emph{embedding}.  Thus, all these terms
    refer to the same process.} of data in a low dimensional space, or perform
  clustering} on the original samples.  In the following we first introduce the Laplacian
matrix and then present a small number of algorithms designed around it.


\section{Intuition}

This section is meant to give an intuitive introduction into the Laplacian matrix,
Laplacian eigenmaps, and spectral clustering.  It is not necessary to understand the
remainder of the lecture notes but hopefully makes it easier.  If you are short on time
and rich in math and machine learning background, you might prefer to skip it.

\markText{The Laplacian matrix can} be used to model heat diffusion in a graph.  Its
theory can thus \markText{be understood intuitively with the help of the heat diffusion
  analogy.}

\subsection{Heat diffusion analogy of Laplacian eigenmaps}

\begin{figure}[htbp!]
\centering
\includegraphics[width=\textwidth]{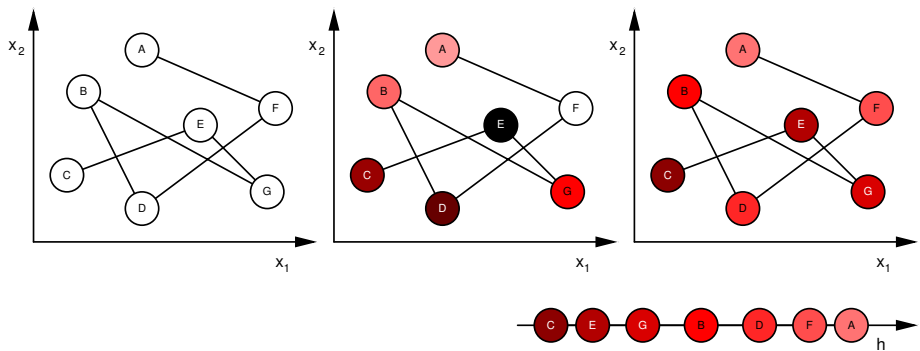}

\caption{\label{Fig:InANutshell-LEM} Heat diffusion analogy of the Laplacian eigenmaps algorithm.}
\end{figure}

First consider a very simple heat diffusion analogy for nonlinear dimensionality
reduction from 2D to 1D with the Laplacian eigenmap algorithm.
\markText{Figure~\ref{Fig:InANutshell-LEM} (left) shows seven points in 2D,} labeled A
through G.  Their position might not be very meaningful but \markText{we assume that we
  have some similarity function that induces relationships between these points.  This
  results in a simple undirected graph} with seven nodes and six edges in this example.
We see already that the graph is a simple linear graph, a chain, but in high dimensions
with many more nodes and a slightly more complicated structure, this might not be so
obvious anymore.

The heat diffusion analogy now says that \markText{nodes are considered heat reservoirs
  and heat can diffuse from one node to neighboring nodes via the edges, but no heat gets
  lost or added.}  So, let us randomly initialize the nodes with arbitrary temperatures,
Figure~\ref{Fig:InANutshell-LEM} (middle).  What happens if we wait?  Well, it is obvious
that heat diffuses from warmer to colder nodes until temperature has balanced out
completely.  It is also obvious that local temperature differences balance out quickly,
while global temperature differences between distant nodes (distant in terms of the graph
connectivity) take more time to balance out.  So \markText{if one measures the
  temperatures quite late in the process, one finds a distribution like the one shown in
  Figure~\ref{Fig:InANutshell-LEM} (right).}  One end of the chain is slightly warmer
than the other end, and from one end to the other there is a monotonic decrease of
temperature.  This is interesting, because \markText{if one now plots the seven points}
again, but now \markText{in a 1D space according to their temperature, one gets the plot
  in Figure~\ref{Fig:InANutshell-LEM} (bottom right).  The points are nicely ordered by
  their position in the linear graph.}  This is much better for visualization and
interpretation and possibly further processing of the points, since the position in space
now reflects similarity relations well.  (The details of the spacing reveal a flattening
of the temperature profile towards the ends, an effect that takes more effort to
understand intuitively and is beyond the scope of this introduction.)

\markText{This is essentially how the Laplacian eigenmaps algorithm works, except that
  one} does not really use heat diffusion but \markText{finds the resulting heat
  distribution analytically} in a more efficient and robust way.  It is also possible to
map the points into a 2D or even higher-dimensional space by taking more than one heat
diffusion mode into account.

\subsection{Heat diffusion analogy of spectral clustering}

\begin{figure}[htbp!]
\centering
\includegraphics[width=\textwidth]{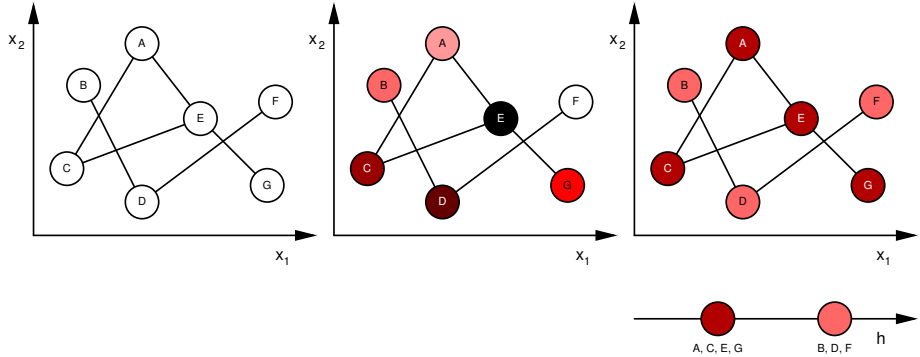}

\caption{\label{Fig:InANutshell-SC} Heat diffusion analogy of spectral clustering.}
\end{figure}

For a heat diffusion analogy of spectral clustering consider a different connectivity of
the graph, like the one shown \markText{in Figure~\ref{Fig:InANutshell-SC} (left).}  The
difference to the example above is that now \markText{the graph has two disconnected
  subgraphs.}  No heat can diffuse from one subgraph to the other.  \markText{If one
  waits long enough, the temperature within each subgraph has completely balanced out,
  but the two subgraphs have different temperature,} because there is no edge between
them, \markText{Figure~\ref{Fig:InANutshell-SC} (right).  If one now plots the seven
  points in a 1D space according to} their \markText{temperature,
  Figure~\ref{Fig:InANutshell-SC} (bottom right),} all points of one subgraph cluster at
one value and the points of the other subgraph cluster at another value.  Thus, in this
space \markText{separating the two subgraphs is trivial.}

\markText{This is essentially how spectral clustering works.}  In real data the clusters,
i.e.\ subgraphs, might not be completely disconnected, but with some tricks one can also
deal with that.

The graphs \markText{in Figures~\ref{Fig:InANutshell-LEM} and~\ref{Fig:InANutshell-SC}}
are drawn in a way that \markText{the position of the nodes} actually \markText{has no
  meaning at all.}  This is to emphasize that the edges are the only thing that matters
for the result of Laplacian eigenmaps and spectral clustering.  \markText{In real world
  examples,} however, spatial proximity often plays an important role and
\markText{edges are preferably inserted between neighboring data points.}

\subsection{Heat diffusion equation for connected heat reservoirs}

How can we model heat diffusion mathematically, and how can we figure out the relevant
temperature distributions analytically?  Heat diffusion is a continuous process, so we
need a differential equation (DE) for it.  Since we consider heat diffusion between a
discrete set of heat reservoirs rather than on a continuous medium, the DE is a system of
ordinary DEs and not a partial DE.  It is linear, e.g.\ if you have twice as much heat,
diffusion will be twice as strong.  And it is homogeneous, because if there is no heat,
then there is no diffusion.  Thus we \markText{consider the following system of ordinary
  linear DEs}
\begin{alignat}{3}
  \markByColor{\markEqnLect} & & \markByColor{\dot{\vec{h}}(t)}&\markByColor{\,=\, - \mat{L}\vec{h}(t)} \label{eq:nutshell:HeatODE} \\
  \markEqnLect &\Longleftrightarrow & \dot{h}_1(t)&\,=\, - L_{11}h_1(t) -L_{12}h_2(t) -L_{13}h_3(t) \\
  & & \wedge\quad \dot{h}_2(t)&\,=\, - L_{21}h_1(t) -L_{22}h_2(t) -L_{23}h_3(t) \\
  & & \wedge\quad \dot{h}_3(t)&\,=\, - L_{31}h_1(t) -L_{32}h_2(t) -L_{33}h_3(t)
\end{alignat}
spelled out for three heat reservoirs, \markText{where $\vec{h}(t)$ is} a nonnegative
vector representing \markText{the temperatures of the nodes} as a function of time.
\markText{$\mat{L}$ is a matrix representing the heat diffusion} between the nodes, and
it will be explained in a moment.

Readers not so familiar with differential equations might find it easier to consider the temporally discretized version of it,
\begin{alignat}{3}
  & & \frac{\vec{h}(t+\Delta t)-\vec{h}(t)}{\Delta t} &\,=\, \dot{\vec{h}}(t) \quad \text{(for $\Delta t\to0$)} \\
  & & &\stackref{\ref{eq:nutshell:HeatODE}}{\,=\,} - \mat{L}\vec{h}(t) \\
  &\Longleftrightarrow & \vec{h}(t+\Delta t)&\,=\, \vec{h}(t) - \Delta t \, \mat{L}\vec{h}(t) \\
  & & &\,=\, (\mat{I} - \Delta t \, \mat{L})\vec{h}(t) \label{eq:nutshell:HeatMap} \\
  &\Longleftrightarrow & h_1(t+\Delta t)&\,=\, h_1(t) - \Delta t ( L_{11}h_1(t) + L_{12}h_2(t) + L_{13}h_3(t)) \\
  & & \wedge\quad h_2(t+\Delta t)&\,=\, h_2(t) - \Delta t ( L_{21}h_1(t) + L_{22}h_2(t) + L_{23}h_3(t)) \\
  & & \wedge\quad h_3(t+\Delta t)&\,=\, h_3(t) - \Delta t ( L_{31}h_1(t) + L_{32}h_2(t) + L_{33}h_3(t))
\end{alignat}
which is an approximation of the differential equation $\dot{\vec{h}}(t) = - \mat{L}\vec{h}(t)$,
which is exact for $\Delta t\to0$.

\subsection{Laplacian matrix}\label{sec:LaplacianMatrix}

In either case, it is clear that $\mat{L}$ is responsible for any change of $\vec{h}$ and
that \markText{the physics of the heat diffusion process imposes constraints on
  $\mat{L}$.}  If $\mat{L}=\mat{0}$ then $\vec{h}(t)$ is constant, which would correspond
to three disconnected nodes (= heat reservoirs) that do not exchange any heat.  A
negative $L_{\idxx\iidxx}$ indicates that $h_\idxx$ increases proportional to $h_\iidxx$ with factor
$-L_{\idxx\iidxx}$.  A positive $L_{\idxx\iidxx}$ indicates that $h_\idxx$ decreases proportional to $h_\iidxx$ with
factor $-L_{\idxx\iidxx}$.

We want that no heat gets lost or added to the system, thus
$\sum_\idxx L_{\idxx\iidxx} = 0$ must be fulfilled, as one can easily verify by setting
$\dot{h}_1(t)+\dot{h}_2(t)+\dot{h}_3(t)=0$ or
$h_1(t+\Delta t)+h_2(t+\Delta t)+h_3(t+\Delta t)=$~const for any values of
$h_1(t), h_2(t)$, and $h_3(t)$.  Since the heat one node gains must come from some other
nodes, one can say that $-L_{\idxx\iidxx}h_\iidxx$ (with negative $L_{\idxx\iidxx}$)
indicates the amount of heat node~$\idxx$ gains from node~$\iidxx$ for $\idxx \ne \iidxx$.
The term $-L_{\iidxx\iidxx}h_\iidxx$ (with positive $L_{\iidxx\iidxx}$) indicates how
much heat node $\iidxx$ looses to the other nodes.

If we consider the situation that all three nodes are connected and one node, say Node~1,
is hot and the other two nodes are absolutely freezing, i.e.\ $h_2=h_3=0$ (Kelvin not
Celsius) then initially only $L_{11}, L_{21}$, and $L_{31}$ matter.  It is intuitively
clear that in this situation heat diffuses from Node~1 to Nodes~2 and~3, i.e.\ $h_1$
decreases and $h_2$ as well as $h_3$ increase proportionally to $h_1$.  This implies
$0 < L_{11}$, indicating that Node~1 looses heat, and $L_{21}, L_{31} < 0$, indicating
that Nodes~2 and~3 gain heat from Node~1.  If a connection would be absent, e.g.\ between
Nodes~2 and~1, then no heat diffuses between these two nodes and the corresponding entry
is zero, $L_{21}=0$.  If a node, let say Node~1, is not connected to any other node, then
it cannot gain or loose heat at all, resulting in $L_{11}=0$.  Thus, by symmetry
arguments we have $0 \le L_{\idxx\idxx}$ and $L_{\idxx\iidxx} \le 0 \ \forall \iidxx\ne\idxx$.

Finally, it is clear that if two different nodes~$\idxx$ and~$\iidxx$ have same temperature,
$h_\idxx=h_\iidxx$, then the heat $-L_{\idxx\iidxx}h_\iidxx$ diffusing from node~$\iidxx$ to node~$\idxx$ equals the heat
$-L_{\iidxx\idxx}h_\idxx$ diffusing from node~$\idxx$ to node~$\iidxx$, because otherwise one node would
spontaneously become warmer and the other cooler, which would allow us to build a
perpetual mobile.  This implies $L_{\idxx\iidxx}=L_{\iidxx\idxx}$.  Please notice here that if two
connected nodes have same temperature, it does not mean that no heat diffuses from one to
the other, it only means that the heat flows cancel out each other.

If we summarize the insights above \markText{we find that}
\markByColor{
  \begin{alignat}{3}
    & & L_{\idxx\iidxx}&\,=\, L_{\iidxx\idxx} & &\quad\text{($\mat{L}$ is symmetric)} \label{eq:nutshell:LSymmetry} \\
    & & \sum_\idxx L_{\idxx\iidxx} \stackref{\ref{eq:nutshell:LSymmetry}}{\,=\,} \sum_\iidxx L_{\idxx\iidxx} &\,=\, 0   & &\quad\text{(rows and columns add up to zero)} \label{eq:nutshell:LSumZero} \\
    & & L_{\idxx\idxx}&\,\ge 0 & &\quad\text{(diagonal elements are non-negative)} \\
    & & L_{\idxx\iidxx}&\,\le 0 \quad \forall \iidxx\ne\idxx& &\quad\text{(off-diagonal elements are non-positive)} \label{eq:nutshell:LNegativeOffDiagonal}
  \end{alignat}
}

An example of a matrix with all these properties is
\begin{alignat}{3}
  \mat{L} \,=\,
  \left(
    \begin{array}{rrrr}
       0.2 & -0.2 &  0\phantom{.0} \\
      -0.2 &  1.0 & -0.8 \\ 
       0\phantom{.0} & -0.8 &  0.8
    \end{array}
  \right)
\end{alignat}
The corresponding graph is shown in Figure~\ref{FIG_simple_graph}.

\subsection{Solution of the heat diffusion equation}

\markText{Assume the eigenvectors $\eVecL_\idxu$ and eigenvalues $\eValL_\idxu$ of the Laplacian
  matrix are known} with
\markByColor{
  \begin{alignat}{3}
    \mat{L}\eVecL_\idxu \,=\, \eValL_\idxu\eVecL_\idxu \label{eq:nutshell:EVEL}
  \end{alignat}
}\markText{and ordered such that $\eValL_1 \le \eValL_2 \le ... \le \eValL_\Idxu$.}  It
turns out that \markText{all eigenvalues are non-negative and}
from~(\ref{eq:nutshell:LSumZero}) follows directly that \markText{one can chose
  $\eVecL_1 = (1,1,...,1)^T$} (usually normalized to norm one by convention)
\markText{with $\eValL_1=0$} as the first eigenvector and -value.

For the discretized version of the differential equation it is interesting to see that
\begin{alignat}{3}
  & & (\ref{eq:nutshell:HeatMap}) \,=\, \underbrace{(\mat{I}-\Delta t \mat{L})}_{=:\,\mat{P}} \eVecL_\idxu&\,=\, \mat{I} \eVecL_\idxu -\Delta t \mat{L} \eVecL_\idxu \label{eq:nutshell:defP} \\
  & & &\stackref{\ref{eq:nutshell:EVEL}}{\,=\,} \eVecL_\idxu - \Delta t \eValL_\idxu \eVecL_\idxu \\
  & & &\,=\, \underbrace{(1-\Delta t \eValL_\idxu)}_{=:\,\xi_\idxu}\eVecL_\idxu  \label{eq:nutshell:EVEP}
\end{alignat}
Thus the $\eVecL_\idxu$ are also eigenvectors of $\mat{P}$ but with eigenvalues
$\xi_\idxu=(1-\Delta t \eValL_\idxu)$ with $1 = \xi_1 \ge \xi_2 \ge ... \ge \xi_\Idxu > 0$ for small enough $\Delta t$.

\markText{Because the Laplacian matrix is symmetric and real, the set of eigenvectors is
  complete, and} any initial temperature vector $\vec{h}(t=0)$ can be written as a linear
combination of the eigenvectors
\begin{alignat}{3}
  \vec{h}(t=0) \,=\, \sum_\idxu \weights_\idxu \eVecL_\idxu \label{eq:nutshell:defH}
\end{alignat}
with some appropriate prefactors $\weights_\idxu$.

From the theory of systems of homogeneous linear differential equations we know that
\markText{the general solution of~(\ref{eq:nutshell:HeatODE})} for this $\vec{h}(t=0)$
\markText{is}
\markByColor{
  \begin{alignat}{3}
    \markEqnLect & & \vec{h}(t)&\,=\, \sum_\idxu \weights_\idxu \exp(-\eValL_\idxu t) \eVecL_\idxu
  \end{alignat}
}

For those who prefer the discretized version of the differential equation one can show that 
\begin{alignat}{3}
  & & \vec{h}(t=N \Delta t)&\stackref{\ref{eq:nutshell:HeatMap},\ref{eq:nutshell:defP}}{\,=\,} \mat{P}^N\vec{h}(0) \\
  & & &\stackref{\ref{eq:nutshell:defH}}{\,=\,} \mat{P}^N\sum_\idxu \weights_\idxu \eVecL_\idxu \\
  & & &\,=\, \sum_\idxu \weights_\idxu \mat{P}^N\eVecL_\idxu \\
  & & &\stackref{\ref{eq:nutshell:EVEP}}{\,=\,} \sum_\idxu \weights_\idxu \xi_\idxu^N\eVecL_\idxu
\end{alignat}

In either case, if one waits long enough, only the first eigenvectors with eigenvalue
$\eValL_\idxu=0$ respectively $\xi_\idxu=1$ will still contribute to $\vec{h}(t)$, and one
can show that if the graph is connected, only the contribution of \markText{$\eVecL_1$
  survives indefinitely long,} because $\exp(-\eValL_1 t)=\exp(-0 t)=1$ and
$\xi_1^N=1^N=1$ for any $t$.  \markText{The last eigenvector fading away is $\eVecL_2$,
  and that is exactly the vector we are interested in for the Laplacian eigenmaps
  algorithm, see Figure~\ref{Fig:InANutshell-LEM} (right).}

\markText{If the graph is disconnected} then it is intuitively clear that each subgraph
balances out its heat over time, but there is no heat exchange between subgraphs.
\markText{The corresponding Laplacian matrix becomes a block matrix} with as many blocks
on the diagonal as there are subgraphs.  In the example above in
Figure~\ref{Fig:InANutshell-SC}, there are two subgraphs, and because of the block
structure of the Laplacian matrix and the fact that rows add up to zero, one can verify
that \markText{the second eigenvector}
$\markByColor{\eVecL_2}=(\sfrac{1}{4},-\sfrac{1}{3},\sfrac{1}{4},-\sfrac{1}{3},\sfrac{1}{4},-\sfrac{1}{3},\sfrac{1}{4})^T$
(usually normalized to norm one by convention) \markText{is constant within each subgraph
  and has eigenvalue $\eValL_2=0$.  This again} reflects the temperature distribution
that remains if one waits for a long time, and that \markText{is exactly the vector we
  are interested in} the spectral clustering algorithm, see
Figure~\ref{Fig:InANutshell-SC} (right).

\markText{In summary, the second eigenvector of the Laplacian matrix provides a nice 1D
  arrangement of the nodes of a similarity graph.}  In practice \markText{one often also
  uses the third and possibly the forth eigenvector to get visualizations in 2D or 3D,}
but that is not so easy to understand with this intuitive explanation.


\section{Formalism}

After the intuitive explanation we \markText{now consider Laplacian eigenmaps and
  spectral clustering} more directly and \markText{more formally.}  For both algorithms
data must first be represented as a graph.  Nodes represent data samples and edges
represent similarities between data samples.  The samples could be anything, e.g.\ words,
persons, or melodies, they need not be vectors in a vector space.  We just need a
non-negative function that measures similarity between two data samples.  And this
function does not even need to be consistent with a metric.  We first introduce some
notions from graph theory and then consider the optimization problem.

\subsection{Simple graphs}\label{sec:SimpleGraph}

\markText{A \emph{graph}}\GWord{Graph} $G = ( \mathbb{V} , \mathbb{E} )$ \markText{is a
  set of \emph{nodes}} (or vertices or
points)\GWord{Knoten}~$\mathbb{V} = \{v_1, ..., v_\Idxx\}$ \markText{and a set of
  \emph{edges}}\GWord{Kanten}~$\mathbb{E} = \{ e_1, ..., e_\IdxE\}$.  \markText{An edge $e_\idxe$
  connects two nodes $v_\idxx$ and $v_\iidxx$} and is therefore defined by a pair of nodes.  Edges
may be \emph{directed}\GWord{gerichtet}, going from node $v_\idxx$ to node $v_\iidxx$, indicated
by $e_\idxe=(v_\idxx,v_\iidxx)$.  Edges may also be \emph{undirected}\GWord{ungerichtet}, in which
case the order of the vertices does not matter and we can write $e_\idxe:=\{v_\idxx,v_\iidxx\}$, where
the curly brackets imply that the order does not matter.  \markText{\emph{Simple
    graphs}}\GWord{einfacher Graph, schlichter Graph} \markText{are undirected graphs
  without loops}\GWord{Schleife, Saline}, which are edges that connect a node with
itself, \markText{and no parallel edges}, which are edges that connect the same pair of
nodes.  \markText{Here we consider mainly simple graphs.}

Further reading: \cite[]{Wikipedia-2017-Graphs}.

\subsection{Matrix representation}\label{sec:MatrixRepresentation}

Graphs can be conveniently represented by real matrices.  \markText{The \emph{adjacency
    matrix}}\GWord{Adjazenzmatrix, Nachbarschaftsmatrix}
\markText{$\mat{A}=(A_{\idxx\iidxx})$ of an undirected graph is} $\Idxx \times \Idxx$ and
\markText{defined as} \markByColor{
  \begin{equation}
  \markEqn A_{\idxx\iidxx} \,:=\,
  \begin{cases}
    1&\text{if }\{v_\idxx,v_\iidxx\}\in\mathbb{E}\\
    0&\text{otherwise}
  \end{cases}
  \label{eq:adjacency}
\end{equation}
}i.e.\ it has a one in entry $A_{\idxx\iidxx}$ if and only if nodes $v_\idxx$ and $v_\iidxx$ are connected
with each other.  Matrix \markText{$\mat{A}$ is naturally symmetric}, since the edges are
not directed.

\markText{The \emph{degree matrix}}\GWord{Gradmatrix} \markText{$\mat{D}=(D_{\idxx\iidxx})$} of an
undirected graph \markText{is a diagonal matrix, where the diagonal entries $D_{\idxx\idxx}$
  indicate the number of edges connected to node $v_\idxx$.}

In context of the Laplacian matrix, \markText{we generalize these definitions to weighted
  graphs}, where the edges are labeled with a real (positive) number indicating their
weight $W_{\idxx\iidxx}$.  If one simply replaces the $1$ values in\ignore{~(\ref{eq:incidence})
  and}~(\ref{eq:adjacency}) by these weights, then \markText{$\mat{A}$ becomes} the
(edge) \markText{weight matrix $\mat{W}$, and the weighted degree matrix
  $\mat{D}=(D_{\idxx\iidxx})$ gets the sum over all weights of the edges converging on a node in
  their diagonal entries.}
\begin{alignat}{3}
  \markEqn & & D_{\idxx\idxx}&\,:=\, \sum_\iidxx W_{\idxx\iidxx} \,=\, \sum_\iidxx W_{\iidxx\idxx}
  \label{eq:D_elem_def}
\end{alignat}

\begin{figure}[htbp!]
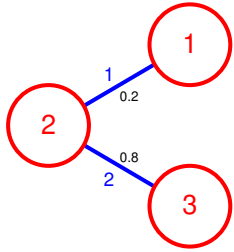

\centering
\hspace*{\fill}
\includegraphics[width=0.25\textwidth]{./Figures/SimpleUndirectedGraph3Nodes}
\ignore{\hspace*{\fill}
\includegraphics[width=0.25\textwidth]{./Figures/SimpleDirectedGraph3Nodes}}
\hspace*{\fill}

\caption{\label{FIG_simple_graph} Example of a simple, weighted, undirected\ignore{ (left) and directed (right)} graph.
Edges are numbered in blue, their weights are shown in black.  Weights do not need to add up to one, like here for Node~2.}
\end{figure}

\LNTODO{Avoid 0.2 and 0.8 as weights in the figure, since they add up to 1, which is not necessary.}

Figure~\ref{FIG_simple_graph} shows\ignore{ two versions of} a simple weighted graph\ignore{, directed on the left, and undirected on the right}.
\ignore{The weighted incidence matrix of the directed graph is
\begin{alignat}{3}
  \mat{B} \,=\,
  \left(
    \begin{array}{ccc}
      & e_1 & e_2 \\ 
      v_1 & +0.2 &  0   \\
      v_2 & -0.2 & -0.8 \\ 
      v_3 &  0   & +0.8 
    \end{array}
  \right)
\end{alignat}}
The weighted adjacency matrix, or weight matrix, of the undirected graph is
\begin{alignat}{3}
  \mat{W} \,=\,
  \left(
    \begin{array}{llll}
      & v_1 & v_2 & v_3 \\ 
      v_1 &  0   &  0.2 & 0   \\
      v_2 &  0.2 &  0   & 0.8 \\ 
      v_3 &  0   &  0.8 & 0
    \end{array}
  \right)
\end{alignat}
The weighted degree matrix of the undirected graph is
\begin{alignat}{3}
  \mat{D} \,=\,
  \left(
    \begin{array}{llll}
      & v_1 & v_2 & v_3 \\ 
      v_1 &  0.2 &  0   & 0   \\
      v_2 &  0   &  1.0 & 0   \\ 
      v_3 &  0   &  0   & 0.8
    \end{array}
  \right)
\end{alignat}

\markText{The \emph{Laplacian matrix}}\GWord{Laplace-Matrix} \markText{$\mat{L}$ is} defined as the
difference between weighted degree matrix $\mat{D}$ and weight matrix $\mat{W}$
\markByColor{
  \begin{equation}
    \markEqn \mat{L} = \mat{D} - \mat{W}. \label{EQ_def_laplacian}
  \end{equation}
}It is easy to verify that it has all the properties
(\ref{eq:nutshell:LSymmetry}--\ref{eq:nutshell:LNegativeOffDiagonal}) derived in
Section~\ref{sec:LaplacianMatrix} from the heat diffusion analogy.

The Laplacian matrix for the example above is
\begin{alignat}{3}
  \mat{L} \,=\,
  \left(
    \begin{array}{rrrr}
       0.2 & -0.2 &  0\phantom{.0} \\
      -0.2 &  1.0 & -0.8 \\ 
       0\phantom{.0} & -0.8 &  0.8
    \end{array}
  \right)
  \label{eq:example:L}
\end{alignat}

\subsection{Optimization problem}

\markText{The objective of Laplacian eigenmaps as well as spectral clustering is to}
assign similar values to similar nodes, i.e.\ strongly connected nodes, and dissimilar
values to nodes that are not similar.  This is a non-trivial operation, since similarity
is a property of a pair of nodes, or an edge, while value is a property of a single node.
It is not guaranteed that there is a good solution at all.  Consider, for instance, three
nodes $A$, $B$, and $C$.  If $A$ and $B$ are very similar as well as $B$ and $C$, but $A$
and $C$ are very dissimilar, then there are no values that could reflect that.  However,
reasonable similarity measures usually do not lead to such conflicts, definitely not
those inducing a proper metric.  In any case, the objective is to \markByColor{
  \begin{alignat}{4}
    \markEqn& & \text{minimize}&\quad  & &\hspace{-5ex}\frac{1}{2} \sum_{\idxx\iidxx} (\eEleL_\idxx - \eEleL_\iidxx)^2 W_{\idxx\iidxx} \label{eq:ObjectiveMinyyW} \\
    \markEqnLect& & \text{subject to}& & \vec{1}^T\eVecL&\,=\, 0 & &\text{\hspace{-5ex}(zero mean)} \label{eq:ConstraintZeroMean} \\
    \markEqnLect& & \text{and}& & \eVecL^T \eVecL&\,=\, 1 & &\text{\hspace{-5ex}(unit variance)} \label{eq:ConstraintUnitVariance} \\
    \markEqn& & \text{or subject to}& & \vec{1}^T \mat{D} \eVecL&\,=\, 0 & &\text{\hspace{-5ex}(weighted zero mean)} \label{eq:ConstraintWeightedZeroMean} \\
    \markEqn& & \text{and}& & \eVecL^T \mat{D} \eVecL&\,=\, 1 & &\text{\hspace{-5ex}(weighted unit variance)} \label{eq:ConstraintWeightedUnitVariance}
  \end{alignat}
}with $\eVecL=(\eEleL_1,\eEleL_2,...,\eEleL_\Idxx)^T$ and $\vec{1}=(1,1,1,...,1)^T$
indicating the one-vector.  Objective~(\ref{eq:ObjectiveMinyyW}) favors solutions where
strongly connected nodes with a large edge weight $W_{\idxx\iidxx}$ have similar values $\eEleL_\idxx$
and $\eEleL_\iidxx$.  Constraints~(\ref{eq:ConstraintZeroMean})
and~(\ref{eq:ConstraintUnitVariance}) in conjunction avoid the trivial constant solution,
which implicitly guarantees that nodes that are not similar get assigned dissimilar
values.  Constraints~(\ref{eq:ConstraintWeightedZeroMean})
and~(\ref{eq:ConstraintWeightedUnitVariance}) have the same function but imply some
normalization, see Section~\ref{sec:RoleOfNormalization}.

If we need more than one solution in order to map the nodes into a higher dimensional
space, we add a subscript index to $\eVecL$ and solve the same optimization problem
multiple times subject to the additional constraint
\begin{alignat}{4}
  \markEqnLect& & & &             \eVecL_\iidxu^T         \eVecL_\idxu&\,=\, 0 \quad \forall \iidxu < \idxu & & \quad \text{(decorrelation to previous solutions)} \label{eq:ConstraintUnitVariance2} \\
  \markEqn& & \text{or} \quad & & \eVecL_\iidxu^T \mat{D} \eVecL_\idxu&\,=\, 0 \quad \forall \iidxu < \idxu & & \quad \text{(decorrelation to previous solutions)} \label{eq:ConstraintWeightedUnitVariance2}
\end{alignat}
for the second and later solutions $\eVecL_\idxu$ to make them different (orthogonal) to the previous solutions $\eVecL_\iidxu$.

\subsection{Associated eigenvalue problem}\label{sec:AssociatedEigenvalueProblem}

It is known that \markText{the normalized eigenvectors $\eVecL_\idxu$ of the ordinary
  eigenvalue equation} \markByColor{
  \begin{equation}
    \markEqnLect \mat{L}\eVecL_\idxu=\eValL_\idxu\eVecL_\idxu \label{eq:OrdinaryEVE}
  \end{equation}
}
  ordered by increasing eigenvalues $\eValL_\idxu$ \markText{solve the optimization problem}
\markByColor{
  \begin{alignat}{4}
  \markEqnLect& & \text{minimize}&\quad  & \eVecL_\idxu^T \mat{L} \eVecL_\idxu&\,=\, \frac{1}{2} \sum_{\idxx\iidxx} (\eEleL_{\idxu,\idxx} - \eEleL_{\idxu,\iidxx})^2 W_{\idxx\iidxx} \label{eq:minwLw} \\
  \markEqnLect& & \text{subject to}& & \eVecL_\idxu^T \eVecL_\idxu&\,=\, 1 & &\text{\hspace{-10ex}(unit norm)}  \label{eq:normalizedI} \\
  \markEqnLect& & \text{and}& & \eVecL_\iidxu^T \eVecL_\idxu&\,=\, 0 \quad\forall \iidxu<\idxu & &\text{\hspace{-10ex}(order and orthogonality)} \label{eq:orderDecorrI}
\end{alignat}
}where constraint~(\ref{eq:orderDecorrI}) induces an order such that $\eVecL_1$ is the
optimal solution without any orthogonality constraint (only the unit norm constraint),
$\eVecL_2$ is the optimal solution with the additional constraint of being orthogonal to
$\eVecL_1$, $\eVecL_3$ is the optimal solution with the additional constraint of being
orthogonal to $\eVecL_1$ and $\eVecL_2$, etc.  Constraints~(\ref{eq:normalizedI},
\ref{eq:orderDecorrI}) can be combined to
$\eVecL_\iidxu^T \eVecL_\idxu \,=\, \delta_{\iidxu\idxu} \ \forall \iidxu\le\idxu$.
Identity~(\ref{eq:minwLw}) is left to the reader as an exercise.  If one orders the
eigenvalues by ascending rather than descending value, the corresponding eigenvectors
solve the maximization rather than minimization problem.  The rest should be known, for
instance from principal component analysis\ignoreForEBISS{ \cite[]{Wisk-LN-PCA}}.

The zero mean constraint~(\ref{eq:ConstraintZeroMean}) is implicit here.  Since the first
solution $\eVecL_1$ is a scaled version of $\vec{1}$, Constraint~(\ref{eq:orderDecorrI}) with
$\iidxu=1$ is equivalent to~(\ref{eq:ConstraintZeroMean}).  The solutions of interest
thus start with index~2 rather than~1.

Since
\begin{equation}
  \eVecL_\idxu^T \mat{L} \eVecL_\idxu \stackref{\ref{eq:OrdinaryEVE}}{\,=\,} \eVecL_\idxu^T \eValL_\idxu \eVecL_\idxu \,=\, \eValL_\idxu \eVecL_\idxu^T \eVecL_\idxu \stackref{\ref{eq:normalizedI}}{\,=\,} \eValL_\idxu \label{eq:eValEqObjective}
\end{equation}
\markText{the eigenvalues are the optimal values of the objective function.}

In the algorithms below the constraint is usually $\eVecLD^T \mat{D} \eVecLD = 1$ rather
than $\eVecL^T \eVecL = 1$ (we switch here from $\eVecL$ to $\eVecLD$ to indicate
solutions with this weighted normalization).  Thus we note that \markText{the
  appropriately normalized eigenvectors $\eVecLD_\idxu$ of the generalized eigenvalue
  equation} \markByColor{
  \begin{equation}
    \markEqn \mat{L}\eVecLD_\idxu=\eValLD_\idxu\mat{D}\eVecLD_\idxu \label{eq:GeneralizedEVE}
  \end{equation}
}ordered by increasing eigenvalues
$\eValLD_\idxu$ \markText{solve the optimization problem}
\markByColor{
  \begin{alignat}{4}
  \markEqn& & \text{minimize}&\quad  & \eVecLD_\idxu^T \mat{L} \eVecLD_\idxu&\,=\, \frac{1}{2} \sum_{\idxx\iidxx} (\eEleLD_{\idxu,\idxx} - \eEleLD_{\idxu,\iidxx})^2 W_{\idxx\iidxx} \label{eq:minuLu} \\
  \markEqn& & \text{subject to}& & \eVecLD_\idxu^T \mat{D} \eVecLD_\idxu&\,=\, 1 & &\text{\hspace{-10ex}(weighted unit norm)} \label{eq:normalizedD} \\
  \markEqn& & \text{and}& & \eVecLD_\iidxu^T \mat{D} \eVecLD_\idxu&\,=\, 0 \quad\forall \iidxu<\idxu & &\text{\hspace{-10ex}(order and weighted orthogonality)} \label{eq:orderDecorrD}
\end{alignat}
}

The derivation~(\ref{eq:eValEqObjective}) does not hold here, since the eigenvectors must
have weighted unit norm, not standard unit norm.  But still we find analogously
\begin{equation}
  \eVecLD_\idxu^T \mat{L} \eVecLD_\idxu \stackref{\ref{eq:GeneralizedEVE}}{\,=\,} \eVecLD_\idxu^T \eValLD_\idxu \mat{D} \eVecLD_\idxu \,=\, \eValLD_\idxu \eVecLD_\idxu^T \mat{D} \eVecLD_\idxu \stackref{\ref{eq:normalizedD}}{\,=\,} \eValLD_\idxu \label{eq:eValDEqObjective}
\end{equation}

Thus, the eigenvalues are the value of the objective function for the different
eigenvectors.  It is intuitively clear that \markText{eigenvectors with small eigenvalue
  are smooth in the sense that connected nodes tend to have similar values} while
eigenvectors with large eigenvalue are more rugged, i.e.\ connected nodes tend to have
different values.

Further reading: \cite[]{Wikipedia-2017-RayleighQuotient}.

\subsection{The role of the weighted normalization constraint}\label{sec:RoleOfNormalization}

\begin{figure}[htbp!]
\centering
\includegraphics[width=0.5\textwidth]{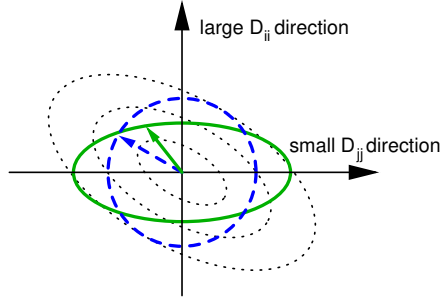}

\caption{\label{Fig:Constraints} Visualization of the role of the constraint on the
  optimization problem.  The dotted ellipses illustrate the quadratic form being
  minimized (\ref{eq:minwLw}, \ref{eq:minuLu}), which is the same for both problems.  The
  blue dashed circle and green solid ellipse illustrate the constraints
  (\ref{eq:normalizedI}) and (\ref{eq:normalizedD}), respectively.  The corresponding 
  arrow indicates the optimal solution, which is the point on the circle or ellipse that
  comes closest to the inner dashed ellipses.}
\end{figure}

What is the difference between the constraints $\eVecL_\idxu^T \eVecL_\idxu = 1$
(\ref{eq:normalizedI}) and $\eVecLD_\idxu^T \mat{D} \eVecLD_\idxu = 1$
(\ref{eq:normalizedD})?  Since $\mat{D}$ is a diagonal matrix, this simply means that in
the constraint the components of the generalized eigenvectors get weighted by
$\sqrt{D_{\idxx\idxx}}$ (\ref{eq:D_elem_def}) (the square root comes from the fact that
in $\eVecLD_\idxu^T \mat{D} \eVecLD_\idxu$ the $D_{\idxx\idxx}$ has to be equally
distributed over the two $\eVecLD_\idxu$).  For the term
$\eEleLD_\idxx D_{\idxx\idxx}\eEleLD_\idxx$ to have the same effect size in the
constraint, a component $\eEleLD_\idxx$ with large $D_{\idxx\idxx}$ must be smaller than
one with a small $D_{\idxx\idxx}$.  This is illustrated in Figure~\ref{Fig:Constraints}
by the green solid ellipse vs the blue dashed circle.  The latter is the set of points
with $\eVecL_\idxu^T \eVecL_\idxu = 1$, the former the set with
$\eVecLD_\idxu^T \mat{D} \eVecLD_\idxu = 1$ with large $D_{\idxx\idxx}$ and small
$D_{\iidxx\iidxx}$.

In the figure it is assumed that the determinant of $\mat{D}$ is one.  That does not need
to be the case.  It could be any other positive value, depending on how strong the
weights of the edges are.  However, a consistent scaling of the weights does not change
the solution, so we can assume w.l.o.g.\ that they are scaled such that $|\mat{D}|=1$.

While the constraint differs, the objective function (\ref{eq:minwLw}, \ref{eq:minuLu})
is the same in both cases.  It takes the form of an unisotropic paraboloid, like a
squeezed champagne glass, indicated in Figure~\ref{Fig:Constraints} by dotted ellipses.
Minimizing it under the constraint means finding the point on the blue circle or green
ellipse that comes closest to the inner ellipses.  To the extent the $D_{\idxx\idxx}$ differ, the
components with larger $D_{\idxx\idxx}$ are favored over components with smaller $D_{\idxx\idxx}$,
because they allow the vector $\eVecLD_\idxu$ to move closer to the origin, where
the true minimum of the objective function with value $0$ lies.

However, this does not mean that all components of $\eVecLD_\idxu$ with large $D_{\idxx\idxx}$ become
larger relative to those with small $D_{\idxx\idxx}$.  That depends also on the objective
function.  But the general tendency is that the change from constraint
$\eVecL_\idxu^T \eVecL_\idxu = 1$ to constraint $\eVecLD_\idxu^T \mat{D} \eVecLD_\idxu = 1$ makes the values of
highly connected nodes (with large $D_{\idxx\idxx}$) larger relative to less connected nodes
(with small $D_{\idxx\idxx}$).

Why might that be useful?  Imagine a square lattice of $7\times 7$ nodes, connected with
their four nearest neighbors with equal edge weights one.  This looks like a pretty good
connectivity to represent the 2D layout of the grid.  Now, imagine in the right half of
the grid, each node is connected to its eight nearest neighbors instead of four.  Both,
the four- as well as the eight-neighbor connectivity, are perfectly fine representations
of the 2D layout.  But because the nodes on the right side have more edges, heat would
diffuse faster and temperature would equalize more quickly, leading to more similar
values, the nodes would move closer together in the embedding.  If one uses constraint
$\eVecLD_\idxu^T \mat{D} \eVecLD_\idxu = 1$ this advantage of the more densely connected
half would be somewhat compensated by scaling up the values, which also leads to larger
differences.  This leads to a value distribution that better reflects the 2D layout and
is less influenced by the different density of connections between left and right half.
\ignore{ This is illustrated in Figure~\ref{Fig:GridExample}.  For the evenly connected
  grid (top) the difference between using the simple constraint
  $\eVecL_\idxu^T \eVecL_\idxu = 1$, i.e.\ the ordinary eigenvalue problem, (left) and
  the weighted constraint $\eVecLD_\idxu^T \mat{D} \eVecLD_\idxu = 1$, i.e.\ the
  generalized eigenvalue problem, (right) is minor.  If anything, the former leads to a
  better map.  However, for the grid with the higher connectivity on the right (bottom),
  the simple constraint leads to quite a collapsed map on the dense side while the
  weighted constraint does less so.  }

It is probably also possible to construct examples where the constraint
$\eVecL_\idxu^T \eVecL_\idxu = 1$ gives more desirable results.  But at least it should
be clear now what the effect of \markText{the constraint
  $\eVecLD_\idxu^T \mat{D} \eVecLD_\idxu = 1$} is, it \markText{somewhat counteracts the
  effect of systematically strong (or weak) connections in a region of the graph.}  This
does not tell much about the effects on a more microscopic level.  But it is clear that
it makes no sense to change the value of a single highly connected node and make it too
different from the values of its neighbors, because that really contributes to a bad
value in the objective function.

\ignore{
  \begin{figure}[htbp!]
\includegraphics[width=0.5\textwidth]{./Images/GridEvenOrdinary}%
\includegraphics[width=0.5\textwidth]{./Images/GridEvenGeneralized}

\includegraphics[width=0.5\textwidth,angle=180]{./Images/GridHalfHalfOrdinary}%
\includegraphics[width=0.5\textwidth,angle=180]{./Images/GridHalfHalfGeneralized}

\caption{\label{Fig:GridExample} $7\times 7$ grid with even connectivity (top) or half
  denser connectivity (bottom) mapped with the Laplacian eigenmaps algorithm using the
  simple constraint, i.e.\ ordinary eigenvalue equation, (left) or the weighted
  constraint, i.e.\ generalized eigenvalue problem, (right).}
\end{figure}
}

\ignore{
  
Another way of looking at this is to consider the symmetric normalized Laplacian matrix.
This effectively converts the generalized eigenvalue problem of $\mat{L}$ and $\mat{D}$
into an ordinary one with just $\mat{\hat{L}}$, so that the constraint
$\eVecLD_\idxu^T \mat{D} \eVecLD_\idxu = 1$ converts back into $\eVecHL^T \eVecHL = 1$ with
$\eVecHL = \mat{\overline{D}} \eVecLD_\idxu$.

If we consider a chain of nine nodes linearly connected with weights that are all
equal $1$, then the Laplacian matrix and the normalized Laplacian matrix are
{\tiny
\begin{alignat}{3}
  \mat{L} \,=\,
  \left(
    \begin{array}{rrrrrrrrr}
      1 &-1 &   &   &   &   &   &   &   \\
     -1 & 2 &-1 &   &   &   &   &   &   \\
        &-1 & 2 &-1 &   &   &   &   &   \\
        &   &-1 & 2 &-1 &   &   &   &   \\
        &   &   &-1 & 2 &-1 &   &   &   \\
        &   &   &   &-1 & 2 &-1 &   &   \\
        &   &   &   &   &-1 & 2 &-1 &   \\
        &   &   &   &   &   &-1 & 2 &-1 \\
        &   &   &   &   &   &   &-1 & 1
    \end{array}
  \right) \quad
  \mat{\hat{L}} \,=\,
  \left(
    \begin{array}{rrrrrrrrrr}
      1 &-\frac{1}{\sqrt{2}} &   &   &   &   &   &   &   \\
     -\frac{1}{\sqrt{2}} & 1 &-\frac{1}{2} &   &   &   &   &   &   \\
        &-\frac{1}{2} & 1 &-\frac{1}{2} &   &   &   &   &   \\
        &   &-\frac{1}{2} & 1 &-\frac{1}{2} &   &   &   &   \\
        &   &   &-\frac{1}{2} & 1 &-\frac{1}{2} &   &   &   \\
        &   &   &   &-\frac{1}{2} & 1 &-\frac{1}{2} &   &   &   \\
        &   &   &   &   &-\frac{1}{2} & 1 &-\frac{1}{2} &   &   \\
        &   &   &   &   &   &-\frac{1}{2} & 1 &-\frac{1}{\sqrt{2}} \\
        &   &   &   &   &   &   &-\frac{1}{\sqrt{2}} & 1
    \end{array}
  \right)
\end{alignat}
}

Now imagine we increase the weight of the edges in the right half of the chain by a factor of
$3$, then the Laplacian matrix and the normalized Laplacian matrix are
{\tiny
\begin{alignat}{3}
  \mat{L} \,=\,
  \left(
    \begin{array}{rrrrrrrrr}
      1 &-1 &   &   &   &   &   &   &   \\
     -1 & 2 &-1 &   &   &   &   &   &   \\
        &-1 & 2 &-1 &   &   &   &   &   \\
        &   &-1 & 2 &-1 &   &   &   &   \\
        &   &   &-1 & 4 &-3 &   &   &   \\
        &   &   &   &-3 & 6 &-3 &   &   \\
        &   &   &   &   &-3 & 6 &-3 &   \\
        &   &   &   &   &   &-3 & 6 &-3 \\
        &   &   &   &   &   &   &-3 & 3
    \end{array}
  \right) \quad
  \mat{\hat{L}} \,=\,
  \left(
    \begin{array}{rrrrrrrrrr}
      1 &-\frac{1}{\sqrt{2}} &   &   &   &   &   &   \\
     -\frac{1}{\sqrt{2}} & 1 &-\frac{1}{2} &   &   &   &   &   \\
        &-\frac{1}{2} & 1 &-\frac{1}{2} &   &   &   &   \\
        &   &-\frac{1}{2} & 1 &-\frac{1}{\sqrt{8}} &   &   &   &   \\
        &   &   &-\frac{1}{\sqrt{8}} & 1 &-\frac{\sqrt{3}}{\sqrt{8}} &   &   &   \\
        &   &   &   &-\frac{\sqrt{3}}{\sqrt{8}} & 1 &-\frac{1}{2} &   &   \\
        &   &   &   &   &-\frac{1}{2} & 1 &-\frac{1}{2} &   \\
        &   &   &   &   &   &-\frac{1}{2} & 1 &-\frac{1}{\sqrt{2}} \\
        &   &   &   &   &   &   &-\frac{1}{\sqrt{2}} & 1
    \end{array}
  \right)
\end{alignat}
}

We see that the symmetric normalized Laplacian matrix changes rather little.  Only at the
transition from strongly to weekly connected node is there a glitch.  So we could expect
that its eigenvectors and values are rather similar.  If we assume that the eigenvectors
$\eVecHL$ remain the same, then we can actually say that if we scale the weights of the
right side by a factor of three, then the solution gets scaled by $1/\sqrt{3}$ on the
right side relative to the left side, since $\eVecHL = \mat{\overline{D}} \eVecLD_\idxu$.

} 

\subsection{Symmetric normalized Laplacian matrix}\label{sec:SymNormLapl}

\markText{For the algorithms below, we consider the eigenvalues and -vectors of the
  generalized eigenvalue equation $\mat{L} \eVecLD_\idxu = \eValLD_\idxu4 \mat{D} \eVecLD_\idxu$.}  Since
most of us are more familiar with the ordinary eigenvalue equation, it is interesting to
note that \markText{one can convert the generalized eigenvalue equation into an ordinary
  one and back again.}  This allows us to transfer what we know about ordinary eigenvalue
equations to the generalized ones.

First assume $D_{\idxx\idxx}\ne0\ \forall \idxx$ ($0\le D_{\idxx\idxx}$ is true in any case) and define
\begin{alignat}{3}
  \markEqnLect& & \vec{d}&\,:=\, (D_{11},...,D_{\Idxx\Idxx})^T \label{eq:d} \\
  \markEqnLect& & \vec{\overline{d}}&\,:=\, (\sqrt{D_{11}},...,\sqrt{D_{\Idxx\Idxx}})^T \label{eq:dOverl} \\
  \markEqnLect& & \vec{\underline{d}}&\,:=\, (1/\sqrt{D_{11}},...,1/\sqrt{D_{\Idxx\Idxx}})^T \label{eq:dUnderl} \\
  \markEqnLect& & \mat{D}&\,:=\, \text{diag}(\vec{d}) \,=\,\mat{D}^T \label{eq:D} \\
  \markByColor{\markEqnLect}& & \markByColor{\mat{\overline{D}}}&\markByColorSpace\markByColor{\,:=\, \text{diag}(\vec{\overline{d}}) \,=\,\mat{\overline{D}}^T} \label{eq:DOverl} \\
  \markByColor{\markEqnLect}& & \markByColor{\matUnderlineD}&\markByColorSpace\markByColor{\,:=\, \text{diag}(\vec{\underline{d}}) \,=\,\matUnderlineD^T} \label{eq:DUnderl}
\end{alignat}
so that, for instance, $\mat{\overline{D}}\mat{\underline{D}}=\mat{\underline{D}}\mat{\overline{D}}=\mat{I}$
and $\mat{\overline{D}}\mat{\overline{D}}=\mat{D}$.

Now we convert the generalized eigenvalue equation into an ordinary one.
\begin{alignat}{3}
  \markByColor{\markEqn}& & \markByColor{\mat{L} \eVecLD_\idxu}&\markByColorSpace\markByColor{\stackrel{!}{\,=\,} \eValLD_\idxu \mat{D} \eVecLD_\idxu} \quad | \ \mat{\underline{D}} \cdot \label{eq:GeneralizedEVE2} \\
  \markEqnLect&\Longleftrightarrow \quad  & \mat{\underline{D}} \mat{L} \underbrace{\mat{\underline{D}}\mat{\overline{D}}}_{=\,\mat{I}} \eVecLD_\idxu&\,=\, \mat{\underline{D}} \eValLD_\idxu \underbrace{\mat{\overline{D}}\mat{\overline{D}}}_{=\,\mat{D}} \eVecLD_\idxu \quad \text{(since $\mat{\underline{D}}$ is invertible)} \\
  \markEqnLect&\Longleftrightarrow \quad  & \underbrace{\mat{\underline{D}} \mat{L} \mat{\underline{D}}}_{=:\,\mat{\hat{L}}} \underbrace{\mat{\overline{D}}\eVecLD_\idxu}_{=:\,\eVecHL_\idxu}&\,=\, \eValLD_\idxu  \underbrace{\mat{\underline{D}} \mat{\overline{D}}}_{=\,\mat{I}} \underbrace{\mat{\overline{D}}\eVecLD_\idxu}_{=:\,\eVecHL_\idxu} \\
  \markByColor{\markEqn}&\markByColorSpace\markByColor{\Longleftrightarrow\quad}& \markByColor{\mat{\hat{L}} \eVecHL_\idxu}&\markByColorSpace\markByColor{\,=\, \eValHL_\idxu \eVecHL_\idxu} \label{eq:OrdinaryEVEHat}
%
\end{alignat}
\markText{with}
\markByColor{
  \begin{alignat}{3}
  \markEqn& & \eVecHL_\idxu&\,=\, \mat{\overline{D}} \eVecLD_\idxu \label{eq:fromGeneralToHat} \\
  \markEqn&\Longleftrightarrow \quad & \eVecLD_\idxu&\,=\, \matUnderlineD \eVecHL_\idxu \label{eq:fromHatToGeneral}
\end{alignat}
}\markText{and the \emph{symmetric normalized Laplacian matrix}}
\markByColor{
  \begin{equation}
  \markEqn \mat{\hat{L}} \,:=\, \matUnderlineD\mat{L}\matUnderlineD \label{EQ_def_laplacian-normalized}
\end{equation}
}Thus, \markText{if and only if $\eVecLD_\idxu$ is an eigenvector of the generalized eigenvalue
  equation with eigenvalue $\eValLD_\idxu$, then $\eVecHL_\idxu$ is an eigenvector of the ordinary
  eigenvalue equation with same eigenvalue $\eValLD_\idxu$.}  It is sometimes helpful to switch
back and forth between these two views.

For the example above we find
\begin{alignat}{3}
  \mat{\hat{L}} \,=\,
  \left(
    \begin{array}{rrrr}
      & \div\sqrt{0.2} & \div\sqrt{1.0} & \div\sqrt{0.8} \\
      & \downarrow\ \  & \downarrow\ \  & \downarrow\ \  \\
      \div\sqrt{0.2} \rightarrow &  0.2 & -0.2 &  0\phantom{.0} \\
      \div\sqrt{1.0} \rightarrow & -0.2 &  1.0 & -0.8 \\ 
      \div\sqrt{0.8} \rightarrow & 0\phantom{.0} & -0.8 &  0.8
    \end{array}
  \right)
  \,=\,
  \left(
    \begin{array}{rrrr}
       1.0 & -\sqrt{0.2} &  0\phantom{.0} \\
      -\sqrt{0.2} &  1.0 & -\sqrt{0.8} \\ 
       0\phantom{.0} & -\sqrt{0.8} &  1.0
    \end{array}
  \right)
  \label{eq:example:hatL}
\end{alignat}
where $\div\sqrt{\cdot}$ indicates multiplication with $\matUnderlineD$ from the left
along the rows and from the right along the columns.  It is easy to see that
$\hat{L}_{\idxx\idxx}=1$ by construction, since
$\matUnderlineD\mat{L}\matUnderlineD = \matUnderlineD(\mat{D}-\mat{W})\matUnderlineD =
(\mat{I}-\matUnderlineD\mat{W}\matUnderlineD)$ and $\matUnderlineD\mat{W}\matUnderlineD$
has only zeroes on the diagonal.  But the rows and columns do not add up to zero anymore.

The objective function related to the eigenvalue equation of the symmetric normalized
Laplacian matrix is
\begin{eqnarray}
  \eVecHL_\idxu^T \mat{\hat{L}} \eVecHL_\idxu&\stackref{\ref{EQ_def_laplacian-normalized}}{=}& \eVecHL_\idxu^T \matUnderlineD\mat{L}\matUnderlineD \eVecHL_\idxu \\
  &=& (\matUnderlineD\eVecHL_\idxu)^T \mat{L}\matUnderlineD \eVecHL_\idxu \quad \text{(since $\matUnderlineD$ is diagonal, thus $\matUnderlineD=\matUnderlineD^T$)} \\
  &\stackref{\ref{eq:minwLw}}{=}& \frac{1}{2} \sum_{\idxx\iidxx} ((\matUnderlineD \eVecHL_\idxu)_\idxx - (\matUnderlineD \eVecHL_\idxu)_\iidxx)^2 W_{\idxx\iidxx} \\
  &\stackref{\ref{eq:DUnderl},\ref{eq:dUnderl}}{=}& \frac{1}{2} \sum_{\idxx\iidxx} \left(\frac{\eVecHLComp_{\idxu,\idxx}}{\sqrt{D_{\idxx\idxx}}} - \frac{\eVecHLComp_{\idxu,\iidxx}}{\sqrt{D_{\iidxx\iidxx}}}\right)^2 W_{\idxx\iidxx} \quad \text{(since $\matUnderlineD$ is diagonal)} \label{eq:objFuncLHat}
\end{eqnarray}  

\subsection{Random walk normalized Laplacian matrix \LNExtraSection}

  
Another possibility to convert the generalized eigenvalue equation into an ordinary one is
simply to multiply (\ref{eq:GeneralizedEVE}) from the left with the inverse of the
weighted degree matrix.
\begin{alignat}{3}
  &\markEqnLect & \mat{L} \eVecLD_\idxu&\stackref{\ref{eq:GeneralizedEVE}}{\,=\,} \eValLD_\idxu \mat{D} \eVecLD_\idxu \quad\quad | \ \mat{D}^{-1} \cdot \\
  &\markEqnLect\Longleftrightarrow \quad & \underbrace{\mat{D}^{-1} \mat{L}}_{=:\,\mat{\hat{L}}^\text{rw}} \eVecLD_\idxu&\,=\, \eValLD_\idxu \eVecLD_\idxu \quad \text{(since $\mat{D}$ is invertible)} \label{EQ_def_laplacian-normalized-rw} \\
  &\markEqnLect\Longleftrightarrow \quad & \mat{\hat{L}}^\text{rw} \eVecLD_\idxu&\,=\, \eValLD_\idxu \eVecLD_\idxu
\end{alignat}
\markText{$\mat{\hat{L}}^\text{rw}:=\mat{D}^{-1}\mat{L}$ is the \emph{random walk
    normalized Laplacian matrix} and has the same eigenvalues and eigenvectors as the
  generalized eigenvalue equation of the Laplacian matrix.}  Its main disadvantage is
that it is non-symmetric.

For the example above we find
\begin{alignat}{3}
  \mat{\hat{L}}^\text{rw} \,=\,
  \left(
    \begin{array}{rrrr}
      \div\, 0.2 \rightarrow &  0.2 & -0.2 &  0\phantom{.0} \\
      \div\, 1.0 \rightarrow & -0.2 &  1.0 & -0.8 \\ 
      \div\, 0.8 \rightarrow & 0\phantom{.0} & -0.8 &  0.8
    \end{array}
  \right)
  \,=\,
  \left(
    \begin{array}{rrrr}
       1.0 & -1.0 &  0\phantom{.0} \\
      -0.2 &  1.0 & -0.8 \\ 
       0\phantom{.0} & -1.0 &  1.0
    \end{array}
  \right)
  \label{eq:example:hatL-rw}
\end{alignat}
where $\div\cdot$ indicates multiplication with $\mat{D}^{-1}$ from the left along the
rows.  Notice that $\hat{L}^\text{rw}_{\idxx\idxx}=1$ and that the rows, but not the columns, add
up to zero.  $\mat{P} \,:=\, \mat{I}-\mat{\hat{L}}^\text{rw}$ is a \emph{right stochastic
  matrix} \cite[]{Wikipedia-2017-StochasticMatrix}, which can be interpreted as a
transition matrix for a random walk between the nodes of the graph.  Therefore the name.
We are not sure how useful this intuition is, since the right stochastic matrix has to be
multiplied from the right, in order to simulate a random walk, but in the eigenvalue
equation $\mat{\hat{L}}^\text{rw}$ is multiplied from the left.

In what follows we focus on $\mat{\hat{L}}$ rather than $\mat{\hat{L}}^\text{rw}$,
because the non-symmetry makes the latter more difficult to deal with.


\LNLecture{2}{2}{}

\subsection{Summary of mathematical properties}\label{sec:MathProp}

The Laplacian matrix appears in a multitude of different algorithms, three of which will be discussed in this lecture:
\emph{Laplacian eigenmaps (LEM)}, \emph{locality preserving projections (LPP)}, and \emph{spectral clustering}.
When using the Laplacian matrix in an algorithm, we are usually interested in its eigenvectors and eigenvalues.
The set of eigenvalues of a matrix is referred to as its \emph{spectrum}.

The Laplacian matrix, its eigenvectors, and its spectrum have the following properties:
\begin{enumerate}
\item\label{item:LSymmetry}
  \markText{$\mat{L}$ and $\mat{\hat{L}}$ are both symmetric (and real).}
  The symmetry of $\mat{L}$ follows directly from equation~(\ref{EQ_def_laplacian})
  since $\mat{D}$ is diagonal and $\mat{W}$ is symmetric.
  The symmetry of $\mat{\hat{L}}$ follows from equation~(\ref{EQ_def_laplacian-normalized})
  and the symmetry of $\mat{L}$.
  See~(\ref{eq:example:L}) and~(\ref{eq:example:hatL}) for the example above.
\item\label{item:orthogonal}
  \markText{$\mat{L}$ and $\mat{\hat{L}}$ each have a complete set of orthogonal eigenvectors} $\eVecL_\idxu$ and $\eVecHL_\idxu$,
  respectively, with real eigenvalues.
  This is true for any real symmetric matrix, see Property~\myProperty{\ref{item:LSymmetry}}.
\item\label{item:LPositiveSemiDefinite}
  \markText{$\mat{L}$ and $\mat{\hat{L}}$ are both positive semi-definite.}
  For $\mat{L}$ this follows directly from (\ref{eq:minwLw}) and the fact that all weights are positive; for $\mat{\hat{L}}$ this follows from equation~(\ref{EQ_def_laplacian-normalized}) and the fact that it holds for $\mat{L}$.
\item\label{item:NonNegativeEVal}
  \markText{$\mat{L}$ and $\mat{\hat{L}}$ have only non-negative eigenvalues.}
  This follows from Property~\myProperty{\ref{item:LPositiveSemiDefinite}}.
  Note, however, that the eigenvalues of $\mat{L}$ and $\mat{\hat{L}}$ may be different.
  We indicate the eigenvalues of $\mat{L}$ by $\eValL_\idxu$ and those of $\mat{\hat{L}}$ by $\eValHL_\idxu$.
\item\label{item:HatLDEquiv}
  \markText{$\mat{\hat{L}}\eVecHL_\idxu=\eValHL_\idxu\eVecHL_\idxu$ and
    $\mat{L}\eVecLD_\idxu=\eValLD_\idxu\mat{D}\eVecLD_\idxu$ have the same set of
    eigenvalues $\eValLD_\idxu$ and their eigenvectors are related by
    $\eVecLD_\idxu=\matUnderlineD \eVecHL_\idxu \Leftrightarrow
    \eVecHL_\idxu=\mat{\overline{D}} \eVecLD_\idxu$,} see Section~\ref{sec:SymNormLapl}.
\item\label{item:DOrthogonal} The generalized eigenvalue equation
  \markText{$\mat{L}\eVecLD_\idxu=\eValLD_\idxu\mat{D}\eVecLD_\idxu$ has only
    non-negative eigenvalues $\eValLD_\idxu$ and a full set of eigenvectors
    $\eVecLD_\idxu$ that are orthogonal with respect to the inner product
    $\eVecLD_\iidxu\mat{D}\eVecLD_\idxu$} for $\iidxu\ne\idxu$.  This follows from
  Properties~\myProperty{\ref{item:orthogonal},\ref{item:NonNegativeEVal}} with
  Property~\myProperty{\ref{item:HatLDEquiv}}, since
  $\forall \iidxu\ne\idxu: 0 \stackrel{\mathProperty{\ref{item:orthogonal}}}{=}
  \eVecHL_\iidxu^T\eVecHL_\idxu \stackref{\ref{eq:fromGeneralToHat}}{=}
  \eVecLD_\iidxu^T\mat{\overline{D}}\mat{\overline{D}}\eVecLD_\idxu =
  \eVecLD_\iidxu\mat{D}\eVecLD_\idxu$.
  
  \ignore{
    In one of the exercises we can see that the example above has the eigenvalues $\eValLD_1=0, \eValLD_2=1, \eValLD_3=2$ 
    and eigenvectors $\eVecLD_1=(1,1,1)^T, \eVecLD_2=(-0.8,0,0.2)^T, \eVecLD_3=(1,-1,1)^T$.  
    With that we can verify that, for instance, 
    $\eVecLD_1\mat{D}\eVecLD_2= 1\cdot 0.2\cdot (-0.8) + 1\cdot 1.0\cdot 0 + 1\cdot 0.8\cdot 0.2 = 0$.
  }
\item\label{item:oneVector}
  \markText{$\vec{1}:=(1,1,...,1)^T$ (the one-vector) 
    is a solution of} the ordinary eigenvalue equation \markText{$\mat{L}\eVecL_\idxu=\eValL_\idxu\eVecL_\idxu$
    as well as} the generalized eigenvalue equation \markText{$\mat{L}\eVecLD_\idxu=\eValLD_\idxu\mat{D}\eVecLD_\idxu$ with eigenvalue~0}.
  This follows directly from the definition of $\mat{L}$~(\ref{EQ_def_laplacian}), since its rows sum up to zero,
  and because the two eigenvalue equations are identical for $\eValL_\idxu=\eValHL_\idxu=0$.
  We chose the appropriately normalized one-vector to be the first eigenvectors
  $\eVecL_1=\vec{1}/\sqrt{\vec{1}^T\vec{1}}$ and $\eVecLD_1=\vec{1}/\sqrt{\vec{1}^T\mat{D}\vec{1}}$ with $\eValL_1=\eValHL_1=0$.
\item\label{item:DOneVector} $\vec{\overline{d}}$, see (\ref{eq:dOverl}), is a solution
  of the ordinary eigenvalue equation
  $\mat{\hat{L}}\eVecHL_\idxu=\eValHL_\idxu\eVecHL_\idxu$ with eigenvalue~0.  This
  follows from Property~\myProperty{\ref{item:oneVector}} and
  equation~(\ref{eq:fromHatToGeneral}) since
  $\matUnderlineD\vec{\overline{d}} = \vec{1}
  \stackrel{\mathProperty{\ref{item:oneVector}}}{=} \eVecLD_1$.  We chose this
  'square-root degree-vector' normalized to norm one to be the first eigenvector
  $\eVecHL_1=\vec{\overline{d}}/\sqrt{\vec{\overline{d}}^T\vec{\overline{d}}}$ with
  $\eValHL_1=0$.

  \ignore{
    For the example above $\vec{\overline{d}} = (\sqrt{0.2},1.0,\sqrt{0.8})$ and it is easy to verify that
    \begin{alignat}{3}
      \mat{\hat{L}}\vec{\overline{d}} \stackref{\ref{eq:example:hatL}}{\,=\,}
      \left(
        \begin{array}{rrrr}
          1.0 & -\sqrt{0.2} &  0\phantom{.0} \\
          -\sqrt{0.2} &  1.0 & -\sqrt{0.8} \\ 
          0\phantom{.0} & -\sqrt{0.8} &  1.0
        \end{array}
      \right)
      \left(
        \begin{array}{r}
          \sqrt{0.2} \\ 1.0 \\ \sqrt{0.8}
        \end{array}
      \right)
      \,=\,
      \left(
        \begin{array}{r}
        0 \\ 0 \\ 0
        \end{array}
      \right)
    \end{alignat}
  }
\item\label{item:indicatorVectors} Property~\myProperty{\ref{item:oneVector}} generalizes
  to several eigenvalues with eigenvalue~0 for disconnected graphs (the proof is left to
  the reader as an exercise).  \markText{If a graph has $\numSubs$ subgraphs} that are
  intrinsically connected but not mutually, \markText{then $\mat{L}$ has $\numSubs$
    orthogonal eigenvectors with eigenvalue~0.}  Each of these eigenvectors has identical
  values within each of the connected subgraphs and possibly different values between
  subgraphs.  Since it is possible to arbitrarily rotate a set of eigenvectors with
  identical eigenvalue and still get a set of eigenvectors, \markText{it is possible to
    chose the eigenvectors with eigenvalue~0 such that each one has the value~1 within a
    subgraph and value~0 on all other nodes.}  Such vectors are \markText{referred to as
    \emph{indicator vectors}} \cite[]{Wikipedia-2016-IndicatorVector}.  These indicator
  vectors can then be normalized to fulfill the convention of normalized eigenvectors.
\item If we do not perform the rotation mentioned in Property~\myProperty{\ref{item:indicatorVectors}} to
  get indicator vectors, but rather choose the first eigenvector to be the one-vector,
  then \markText{all higher eigenvectors of} the ordinary eigenvalue equation
  \markText{$\mat{L}\eVecL_\idxu=\eValL_\idxu\eVecL_\idxu$ have zero mean,} since
  $\forall \idxu\ne1 : 0 \stackrel{\mathProperty{\ref{item:orthogonal}}}{=} \eVecL_1^T\eVecL_\idxu
  \stackrel{\mathProperty{\ref{item:oneVector}}}{\Longleftrightarrow} 0 = \vec{1}^T\eVecL_\idxu = \sum_\iidxx \eEleL_{\idxu \iidxx}$ by
  Properties~\myProperty{\ref{item:orthogonal},\ref{item:oneVector}}.

\item\label{item:weightedZeroMean} Similarly, if the first eigenvector is the one-vector
  \markText{all higher eigenvectors of} the generalized eigenvalue equation
  \markText{$\mat{L}\eVecLD_\idxu=\eValLD_\idxu\mat{D}\eVecLD_\idxu$ have weighted zero
    mean} since
  $\forall \idxu\ne1 : 0 \stackrel{\mathProperty{\ref{item:DOrthogonal}}}{=}
  \eVecLD_1^T\mat{D}\eVecLD_\idxu \stackrel{\mathProperty{\ref{item:oneVector}}}{\Longleftrightarrow}
  0 = \vec{1}^T\mat{D}\eVecLD_\idxu = \sum_\iidxx \eEleLD_{\idxu \iidxx}D_{\iidxx\iidxx}$ by
  Properties~\myProperty{\ref{item:DOrthogonal},\ref{item:oneVector}}.

\item\label{item:solutions} \markText{The eigenvectors are solutions to the optimization
    problems and the eigenvalues are the values that the objective functions assume for
    the optimal solutions,} see Section~\ref{sec:AssociatedEigenvalueProblem}.  Equation
  (\ref{eq:eValEqObjective}) yields $\eVecL_\idxu^T \mat{L} \eVecL_\idxu = \eValL_\idxu$,
  and $\eVecHL_\idxu^T \mat{\hat{L}} \eVecHL_\idxu = \eValHL_\idxu$ holds analogously.
  For the generalized eigenvalue equation, we find (\ref{eq:eValDEqObjective})
  $\eVecLD_\idxu^T \mat{L} \eVecLD_\idxu = \eValLD_\idxu$.

\end{enumerate}

\begin{sidewaystable}[hbtp!]
  \scriptsize
    \renewcommand{\arraystretch}{1.5}
    \centerline{
    \begin{tabular}{|lc|lc|lc|}
      \hline
      & \multicolumn{3}{c|}{Weight matrix: $\mat{W}$} &
      & \\
    \rlap{(\ref{eq:D_elem_def})}    & \multicolumn{3}{c|}{Degree matrix: $\mat{D} : D_{\idxx\iidxx} \,:=\, \delta_{\idxx\iidxx} \sum_\iidxx W_{\idxx\iidxx}$} &
    \rlap{(\ref{eq:DUnderl})}       & $\matUnderlineD := \text{diag}(1/\sqrt{D_{11}},...,1/\sqrt{D_{\Idxx\Idxx}})$ \\
    \rlap{(\ref{EQ_def_laplacian})} & \multicolumn{3}{c|}{Laplacian matrix: $\mat{L} \,=\, \mat{D} - \mat{W}$} &
    \rlap{(\ref{EQ_def_laplacian-normalized})} & Sym.\ norm.\ Lapl.\ matrix: $\mat{\hat{L}} \,:=\, \matUnderlineD\mat{L}\matUnderlineD$ \\
    \rlap{\myProperty{\ref{item:LSymmetry}}} & \multicolumn{3}{c|}{Is symmetric: $\mat{L} \,=\, \mat{L}^T$} &
    \rlap{\myProperty{\ref{item:LSymmetry}}} & Is symmetric: $\mat{\hat{L}} \,=\, \mat{\hat{L}}^T$ \\
    \rlap{\myProperty{\ref{item:LPositiveSemiDefinite}}} & \multicolumn{3}{c|}{Is positive semi-definite: $\vec{x}^T\mat{L}\vec{x} \,\ge\, 0 \ \forall \, \vec{x}$} &
    \rlap{\myProperty{\ref{item:LPositiveSemiDefinite}}} & Is positive semi-definite: $\vec{x}^T\mat{\hat{L}}\vec{x} \,\ge\, 0 \ \forall \, \vec{x}$ \\
    \hline %
          (\ref{eq:OrdinaryEVE})     & Ordinary eigenvalue eq.: $\mat{L}\eVecL_\idxu \,=\, \eValL_\idxu\eVecL_\idxu$ &
          (\ref{eq:GeneralizedEVE})  & Generalized eigenvalue eq.: $\mat{L}\eVecLD_\idxu\,=\,\eValLD_\idxu\mat{D}\eVecLD_\idxu$ &
          (\ref{eq:OrdinaryEVEHat})  & Ordinary eigenvalue eq.: $\mat{\hat{L}} \eVecHL_\idxu \,=\, \eValHL_\idxu \eVecHL_\idxu$ \\
    & Optimization problem: minimize &
    & Optimization problem: minimize &
    & Optimization problem: minimize \\
    \rlap{(\ref{eq:minwLw})}        & $\eVecL_\idxu^T \mat{L} \eVecL_\idxu \,=\, \frac{1}{2} \sum_{\idxx\iidxx} (\eEleL_{\idxu,\idxx} - \eEleL_{\idxu,\iidxx})^2 W_{\idxx\iidxx}$ &
    \rlap{(\ref{eq:minuLu})}        & $\eVecLD_\idxu^T \mat{L} \eVecLD_\idxu \,=\, \frac{1}{2} \sum_{\idxx\iidxx} (\eEleLD_{\idxu,\idxx} - \eEleLD_{\idxu,\iidxx})^2 W_{\idxx\iidxx}$ &
    \rlap{(\ref{eq:objFuncLHat})}   & $\eVecHL_\idxu^T \mat{\hat{L}} \eVecHL_\idxu \,=\, \frac{1}{2} \sum_{\idxx\iidxx} \left(\frac{\eVecHLComp_{\idxu,\idxx}}{\sqrt{D_{\idxx\idxx}}} - \frac{\eVecHLComp_{\idxu,\iidxx}}{\sqrt{D_{\iidxx\iidxx}}}\right)^2 W_{\idxx\iidxx}$ \\
    \rlap{(\ref{eq:normalizedI},\ref{eq:orderDecorrI})} & subject to $\eVecL_\iidxu^T \eVecL_\idxu \,=\, \delta_{\iidxu\idxu} \ \forall \iidxu\le\idxu$ &
    \rlap{(\ref{eq:normalizedD},\ref{eq:orderDecorrD})} & subject to $\eVecLD_\iidxu^T \mat{D} \eVecLD_\idxu \,=\, \delta_{\iidxu\idxu} \ \forall \iidxu\le\idxu$ &
                                                 & subject to $\eVecHL_\iidxu^T \eVecHL_\idxu \,=\, \delta_{\iidxu\idxu} \ \forall \iidxu\le\idxu$ \\
    & Trivial first solution: &
    & Trivial first solution: &
    & Trivial first solution: \\
    \rlap{\myProperty{\ref{item:oneVector}}} & $\eVecL_1\,=\,\vec{1}/\sqrt{\vec{1}^T\vec{1}}$ with $\eValL_1\,=\,0$ &
    \rlap{\myProperty{\ref{item:oneVector}}} & $\eVecLD_1\,=\,\vec{1}/\sqrt{\vec{1}^T\mat{D}\vec{1}}$ with $\eValHL_1\,=\,0$ &
    \rlap{\myProperty{\ref{item:DOneVector}}} & $\eVecHL_1\,=\,\vec{\overline{d}}/\sqrt{\vec{\overline{d}}^T\vec{\overline{d}}}$  with $\eValHL_1\,=\,0$ \\
    & Objective function value: &
    & Objective function value: &
    & Objective function value: \\
    \rlap{\myProperty{\ref{item:solutions}},(\ref{eq:eValEqObjective})} & $\eVecL_\idxu^T \mat{L} \eVecL_\idxu \,=\, \eValL_\idxu$ &
    \rlap{\myProperty{\ref{item:solutions}},(\ref{eq:eValDEqObjective})}& $\eVecLD_\idxu^T \mat{L} \eVecLD_\idxu \,=\, \eValLD_\idxu$ &
    \rlap{\myProperty{\ref{item:solutions}}} & $\eVecHL_\idxu^T \mat{\hat{L}} \eVecHL_\idxu \,=\, \eValHL_\idxu$ \\
    \hline
    & &
    & \multicolumn{3}{c|}{Relation between two solutions:} \\
    & &
    \rlap{\myProperty{\ref{item:HatLDEquiv}},(\ref{eq:GeneralizedEVE2},\ref{eq:OrdinaryEVEHat})} & \multicolumn{3}{c|}{$\eValLD_\idxu \,=\, \eValLD_\idxu$} \\
    & &
    \rlap{\myProperty{\ref{item:HatLDEquiv}},(\ref{eq:fromHatToGeneral})} & \multicolumn{3}{c|}{$\eVecLD_\idxu \,=\, \matUnderlineD \eVecHL_\idxu$} \\
    \hline
    \end{tabular}
    }
  \normalsize
  \caption{\label{tab:overview} Overview over different Laplacian matrices, eigenvalue equations, optimization problems, and solutions.}
\end{sidewaystable}

Further reading: \cite[]{Wikipedia-2017-LaplacianMatrix}.

\section{Algorithms}

\LNTODO{Consider Isomap by Tenenbaum 2000.}

\LNTODO{Consider the generalized framework by Yan et al 2007.}

\subsection{Similarity graphs}\label{section:simGraphs}

The algorithms presented in the following are all based on the properties of the
Laplacian matrix discussed above. In order to take advantage of the Laplacian matrix,
though, any \markText{input data first has to be represented as a graph}, commonly
\markText{referred to as a \emph{similarity graph}}: A simple graph where the nodes
represent individual data samples and edge weights denote the similarity (or distance)
between two connected nodes, i.e.\ data samples. Appropriate similarity metrics depend on
the problem and can be as simple as the Euclidean or Manhattan distance between two
points.

\markText{There are different ways to construct a similarity graph}, depending on the
problem at hand (e.g.\ \citeauthor{BelkinNiyogi-2003}, \citeyear{BelkinNiyogi-2003},
Sec.~2; \citeauthor{HeNiyogi-2004}, \citeyear{HeNiyogi-2004}, Sec.~2.2;
\citeauthor{VonLuxburg-2007}, \citeyear{VonLuxburg-2007}, Sec.~2). Three common methods
are $\epsilon$-neighborhood, $k$-nearest neighbors, and fully connected graphs:

\begin{itemize}
\item \markText{\textbf{$\epsilon$-neighborhood:} Two nodes are connected if the distance
    between them is smaller than} a given threshold \markText{$\epsilon$.} Often
  $\epsilon$ is chosen so small that the distance values within an
  $\epsilon$-neighborhood do not carry much useful information. In this case
  \markText{edges are} often \markText{weighted} binary, i.e., \markText{with $1$ or $0$}
  depending on whether the data samples in question are close enough or not, respectively.
\item \markText{\textbf{$k$-nearest neighbors:} Node $v_\idxx$ is connected to $v_\iidxx$
    if $v_\iidxx$ is among the $k$ nearest neighbors of $v_\idxx$.} Note that
  \markText{this neighborhood relation is not symmetric} and yields a directed graph,
  thus \markText{some cleanup is required.}  To arrive at a simple graph we take each
  unilateral edge that has no mirrored counterpart and either remove it or keep it and
  set it as bilateral. Removal results in a graph where each node has at most $k$
  neighbors (\emph{mutual $k$-nearest neighbor graph}), while setting unilateral edges to
  bilateral results in a graph where each node has at least $k$ neighbors
  (\emph{$k$-nearest neighbor graph}). All \markText{edges are weighted by the similarity
    between the two nodes they connect.}  Binary weighting, as in the preceding method,
  is more dangerous here, because it cannot be guaranteed that connected nodes are close
  to each other.
\item \markText{\textbf{Fully connected:}} To construct a fully connected graph each data
  sample is simply connected to all others.  In this case, using binary weights renders
  the graph entirely meaningless.  A fully connected graph \markText{always requires
    weighting the edges with a similarity function} (e.g.\ a Gaussian similarity function
  for vectorial data
  $w_{\idxx\iidxx} = w_{\iidxx\idxx} = s(\vec{x}_\idxx,\vec{x}_\iidxx) =
  \exp(-||\vec{x}_\idxx-\vec{x}_\iidxx||^2/(2\sigma^2))$ where $\sigma$ defines the
  extent of local neighborhoods).
\end{itemize}

\LNTODO{Discuss how to chose one of the methods, or give an intuition for their
  (dis)advantages.}

\subsection{Laplacian eigenmaps (LEM)}\label{sec:LEM}

\subsubsection{Motivation}

Many algorithms work only on vectorial data and are limited in the dimensionality they
can process efficiently.  This causes problems if one has data that is either not
vectorial, such as text, or too high dimensional, such as images, or both.  If one can
define a similarity function on the data, yielding a scalar similarity value for each
pair of data samples, the Laplacian eigenmaps algorithm can provide a low-dimensional
vectorial embedding of the data that tends to preserve similarity relationships and
allows to apply other algorithms to the data that would not be applicable directly
\cite[]{BelkinNiyogi-2003}.  Laplacian eigenmaps are also very good for a 2- or
3-dimensional visualization of data.

\ignore{
  A common problem in processing real world data is the high dimensionality of the
individual data samples, particularly when considering real world data such as image
sequences.  \markText{Finding a low dimensional representation} of such data is often an
invaluable -- and sometimes even essential -- step of enabling other algorithms to
process it in a reasonable amount of time. It is commonly referred to as
\emph{dimensionality reduction}.

This process works best when applied to data located on an $n$-dimensional
\emph{manifold} embedded within an $m$-dimensional space, where $n \ll m$. Given a
problem instance in the form of a number of $m$-dimensional data points, the Laplacian
eigenmaps algorithm provides such an $n$-dimensional representation. More precisely, the
algorithm computes a projection of each data point onto a set of $n$ orthogonal
dimensions forming a manifold in the original, $m$-dimensional space \cite[]{BelkinNiyogi-2003}.
}

\textbf{Example:} Imagine a drone hovering through the air while equipped with a downward
facing camera. Using the high dimensional pictures from its camera, we could, in theory,
precisely compute the drone's current position and elevation. Unfortunately, the space of
all possible high dimensional images is effectively intractable. Luckily though, we are
merely interested in a small subset of this space, namely only those images the drone's
camera can actually produce in a particular environment. And while each data point of
this vastly smaller subset still is of the original, high dimensionality, it can be fully
described by six dimensions alone: the position and orientation of the drone in 3D space.
Laplacian eigenmaps can be used to find a low dimensional embedding of the images that
still permits extracting positional and orientation information.

\subsubsection{Objective}

\markText{The objective of the Laplacian eigenmaps algorithm is to find an embedding of a
  set of $\Idxx$ data samples} (do not need to be vectors, but there must be a similarity
function) \markText{in a low-dimensional vector space $\{\vec{y}_1, ... \vec{y}_\Idxx\}$
  such that samples with high similarity are close to each other in the embedding.} For
dimensionality $\IdxN=1$, i.e.\ an embedding in only a $1$-dimensional space, this
objective translates into minimizing
\begin{equation}
\frac{1}{2} \sum_{\idxx\iidxx} (y_\idxx - y_\iidxx)^2 W_{\idxx\iidxx} \label{eq:lapl_min}
\end{equation}
where the $y_\idxx$ are the values assigned to the samples and $W_{\idxx\iidxx}$ indicates
the similarity between two samples.  We have already seen above how this optimization
problem is solved by the second eigenvector of the Laplacian matrix, (\ref{eq:minwLw}) or
(\ref{eq:minuLu}) depending on the constraint.  Each additional eigenvector adds one
orthogonal (meaning the values are uncorrelated) dimension to the embedding provided by
the other eigenvectors already.  The quality of the embedding induced by each eigenvector
is given by its associated eigenvalue, which directly relates to the actual value of sum
(\ref{eq:lapl_min}). \markText{The best $\IdxN$-dimensional embedding is} thus
\markText{given by the first $\IdxN$ eigenvectors $\eVecLD_\idxu$ of the Laplacian matrix
  with smallest eigenvalues} (\markText{excluding the first one}).

Please notice that the dimension of the eigenvectors corresponds to the number $\Idxx$ of
data points, because the Laplacian matrix is $\Idxx\times \Idxx$ by construction.  Thus,
if you arrange the first $\IdxN$ eigenvectors as rows in a matrix, this matrix will be
$\IdxN\times \Idxx$ and the column vectors are the data points $\vec{y}_\idxx$ in the
$\IdxN$-dimensional embedding.  For instance, three data samples embedded in a
2-dimensional space with LEM using the ordinary eigenvalue problem (for simplicity) could
yield
\begin{alignat}{3}
  \left(
    \begin{array}{rccc}
      & \vec{y}_1 & \vec{y}_2 & \vec{y}_3 \\
      & \downarrow & \downarrow & \downarrow \\
      \eVecL_2 \rightarrow & -1/\sqrt{2} &  0          &  +1/\sqrt{2} \\
      \eVecL_3 \rightarrow & -1/\sqrt{6} & +2/\sqrt{6} & -1/\sqrt{6} 
    \end{array}
  \right)
\end{alignat}
As usual, we have dropped $\eVecL_1$, because it has equal components throughout, e.g.\ $(1,1,1)^T/\sqrt{3}$;
$\eVecL_2$ and $\eVecL_3$ have zero mean, because they need to be orthogonal to
$\eVecL_1$; and $\eVecL_2$ and $\eVecL_3$ are orthogonal to each other as well.

We now have all the required components to formulate the Laplacian eigenmaps algorithm.

\subsubsection{Algorithm}\label{sec:LEM-Algorithm}

\begin{enumerate}
\item[] \markText{\textbf{Laplacian eigenmaps algorithm}} \cite[]{BelkinNiyogi-2003}
\item Given a set of $\Idxx$ data samples, \markText{construct a similarity graph $G$}
  according to one of the methods described in Section~\ref{section:simGraphs}.
\item \markText{Construct the} $\Idxx\times\Idxx$ weight matrix $\mat{W}$, degree matrix $\mat{D}$
  (\ref{eq:D_elem_def}), and \markText{Laplacian matrix $\mat{L}$}
  (\ref{EQ_def_laplacian}) \markText{for $G$.}
\item \markText{Compute the first $\IdxN+1$ eigenvectors $\eVecLD_\idxu$ of the generalized
    eigenvalue problem} \markByColor{
    \begin{equation}
      \markEqnLect \mat{L} \eVecLD_\idxu = \eValLD_\idxu \mat{D} \eVecLD_\idxu
    \end{equation}
  }\markText{ordered by increasing eigenvalues.}
\item \markText{An $\IdxN$-dimensional representation of data sample $\idxx$ is now given
    by $(\eEleLD_{2,\idxx}, ..., \eEleLD_{\IdxN+1,\idxx})^T$.}
\end{enumerate}

\subsubsection{Sample applications}

\begin{figure}[htbp!]
\centering
\includegraphics[width=\textwidth]{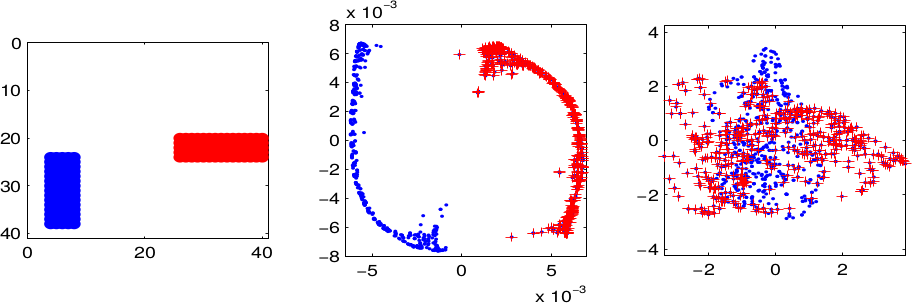}

{\RefSize\RefCLInfoCiteA{}{./Images/BelkinNiyogi-2002-NIPS-LEM-Fig01}}

\caption{\label{fig:LEMBars} Dimensionality reduction of 1000 40$\times$40 images with
  either a vertical or a horizontal bar (plotted together but distinguished by color for
  illustrative purposes, the original is in grayscale).  Left: Two input images
  superimposed, one with a horizontal bar (red) one with a vertical bar (blue).  Middle:
  Result of LEM.  Right: Result of PCA for comparison.}

\end{figure}

\markText{Figure~\ref{fig:LEMBars} shows a toy example of dimensionality reduction of
  1000 images of size 40$\times$40 with either a vertical or a horizontal bar}
\cite[]{BelkinNiyogi-2002}.  One can clearly see how the images with the horizontal bar
are separate from the images with the vertical bar.  It would be interesting to see a
three dimensional Laplacian eigenmap, because presumably the red and blue points would
each form a square manifold representing $x$- and $y$-position.  A projection onto the
first two principal components is shown for comparison.

\begin{figure}[htbp!]
\centering
\includegraphics[width=\textwidth]{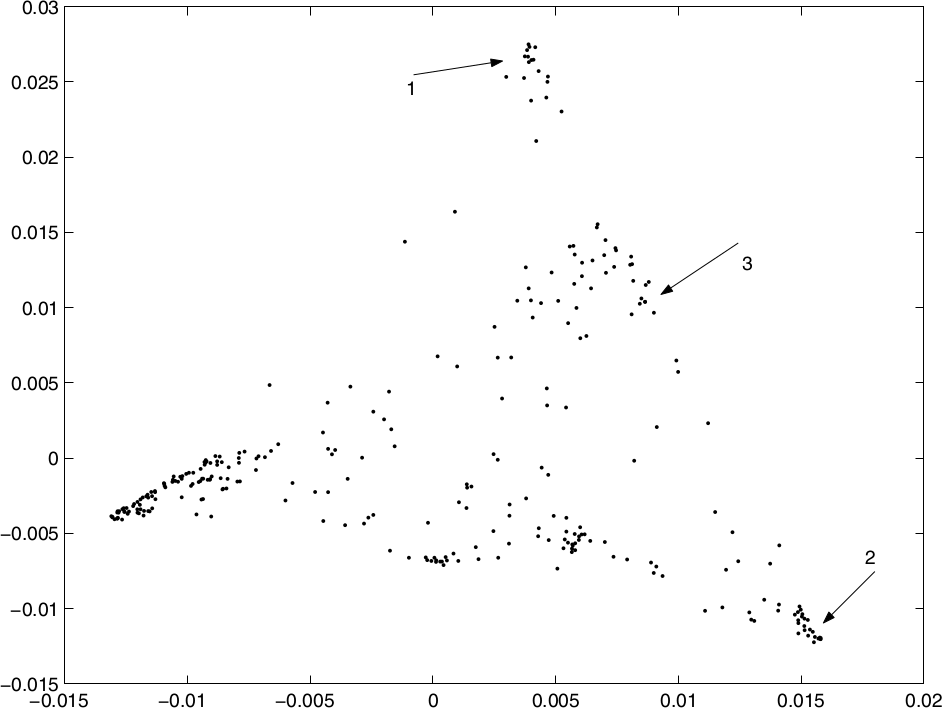}

{\RefSize\RefCLInfoCiteA[Fig.~4]{}{./Images/BelkinNiyogi-2003-NeurComp-LEM}}

\caption{\label{fig:LEMWords} Dimensionality reduction for 300 frequently used words from their word context data.}

\end{figure}

\begin{figure}[htbp!]
\centering
\includegraphics[width=\textwidth]{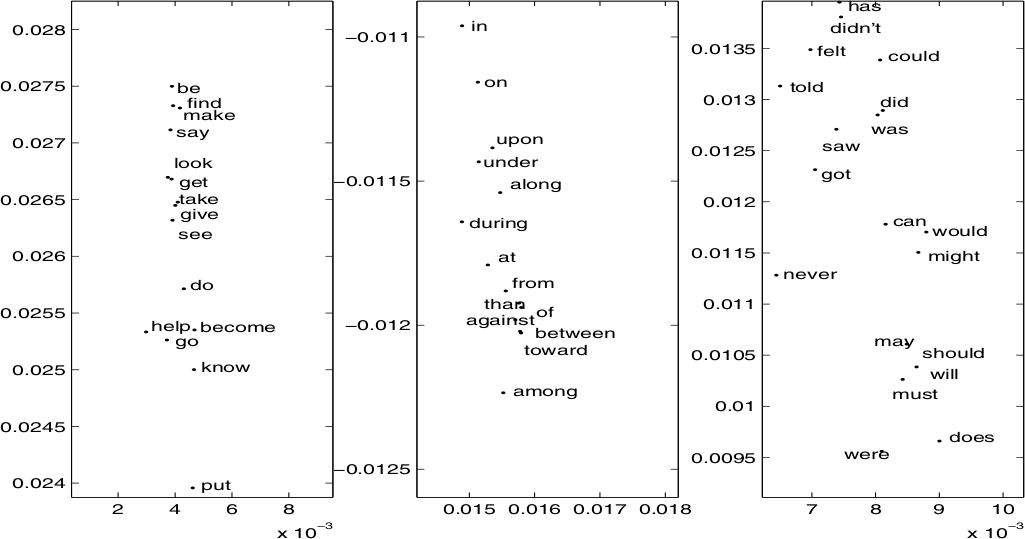}

{\RefSize\RefCLInfoCiteA[Fig.~5]{}{./Images/BelkinNiyogi-2003-NeurComp-LEM}}

\caption{\label{fig:LEMWordsZoomIn} Zoom-in into the three subregions marked in Figure~\ref{fig:LEMWords}.
  Left infinitives, middle prepositions, and right mostly modal and auxiliary verbs.}

\end{figure}

\markText{Figures~\ref{fig:LEMWords} and~\ref{fig:LEMWordsZoomIn} show an application of
  Laplacian eigenmaps to a set of 300 frequently used words} \cite[]{BelkinNiyogi-2003}.
Each word was represented by a 600-dimensional vector indicating how often any of the
other words was found to the left or to the right of the considered word.  Similarity was
defined based on these 600-dimensional vectors.  \markText{Zooming into
Figure~\ref{fig:LEMWordsZoomIn} shows that grammatically closely related words are
grouped together.}

Further reading: \cite[]{BelkinNiyogi-2003}.

\subsection{Locality preserving projections (LPP)}\label{sec:LPP}

\subsubsection{Linear LPP}

\markText{Laplacian eigenmaps} have the disadvantage that they \markText{only provide
  values for the data used during training.}  There is no straight forward way to process
new data.  \markText{This can be changed if} the nodes $v_\idxx$ are data points in
Euclidean space $v_\idxx=\inVec_\idxx\in\mathbb{R}^\inDim$ and \markText{the values of
  the eigenvectors $\eVecLD_\idxu$ are approximated by linear functions in the data
  points} \cite[]{HeNiyogi-2004}.  Since the values of the nodes are now computed with a
linear function rather than assigned freely, new data can be processed by applying the
same linear function.  On the training data the linear function yields the values of the
nodes as follows
\begin{alignat}{3}
  \markEqnLect& & & & \eEleLD_{\idxu,\idxx}&\,=\, \inVec_\idxx^T \funcWVec_\idxu \label{eq:uOfx} \\
  \markByColor{\markEqn}&\Longleftrightarrow\quad & & & \markByColor{\eVecLD_\idxu}&\markByColor{\,=\, \inMat^T \funcWVec_\idxu} \label{eq:UOfX} \\
  \markByColor{\markEqn}&\markByColor{\text{with data}}\quad & & & \markByColor{\inMat}&\markByColor{\,:=\, (\inVec_1,\inVec_2,...,\inVec_\Idxx)}
\end{alignat}
The vectors $\funcWVec_\idxu$ are the variables to be optimized.
Inserting this in~(\ref{eq:minuLu}) and the corresponding constraints~(\ref{eq:normalizedD},\ref{eq:orderDecorrD}) yields
\begin{alignat}{3}
  \markEqn\text{minimize}&\quad  & \eVecLD_\idxu^T \mat{L} \eVecLD_\idxu&\stackref{\ref{eq:UOfX}}{\,=\,}
  \funcWVec_\idxu^T \underbrace{\inMat \mat{L} \inMat^T}_{=:\,\mat{L}'} \funcWVec_\idxu \,=\, \funcWVec_\idxu^T \mat{L}' \funcWVec_\idxu \\
  \markEqnLect\text{subject to}& & 1\,=\, \eVecLD_\idxu^T \mat{D} \eVecLD_\idxu&\stackref{\ref{eq:UOfX}}{\,=\,} \funcWVec_\idxu^T
  \underbrace{\inMat \mat{D} \inMat^T}_{=:\,\mat{D}'} \funcWVec_\idxu \,=\, \funcWVec_\idxu^T \mat{D}' \funcWVec_\idxu  \\
  \markEqnLect\text{and}& & 0 \,=\, \eVecLD_\iidxu^T \mat{D} \eVecLD_\idxu&\stackref{\ref{eq:UOfX}}{\,=\,} \funcWVec_\iidxu^T
  \underbrace{\inMat \mat{D} \inMat^T}_{=:\,\mat{D}'} \funcWVec_\idxu \,=\, \funcWVec_\iidxu^T \mat{D}' \funcWVec_\idxu \quad\forall \iidxu<\idxu 
\end{alignat}

\markText{This optimization problem can} again \markText{be solved} through a generalized
eigenvalue problem, \markText{much like the original one.}  Notice, however, that the
eigenvalues and the approximated eigenvectors $\eVecLD_\idxu$ are not necessarily
identical to those of the original eigenvalue problem, because
$\eVecLD_\idxu\in\mathbb{R}^\Idxu$ is not free but constrained to be a linear function in
the $\inVec_\idxx\in\mathbb{R}^\inDim$.
Notice also that this problem is not of the dimensionality of the number $\Idxx$ of data
points as before but only of the dimension $\inDim$ of the data points, which is usually
much smaller and, consequently makes this approximation more computationally efficient.
For instance, if you have 100 data points in 3D, the problem is 3-dimensional not
100-dimensional as for the LEM algorithm.  The main advantage, however, is that new data
points $\inVec_\iidxx$ can easily be mapped into the low-dimensional space by applying
the linear function $\inVec_\iidxx^T \funcWVec_\idxu$.  Performing Laplacian eigenmaps
with \markText{this} linear approximation \markText{is referred to as \emph{locality
    preserving projections} (LPP).}

\subsubsection{Sample application}

\begin{figure}[htbp!]
\centering
\includegraphics[width=\textwidth]{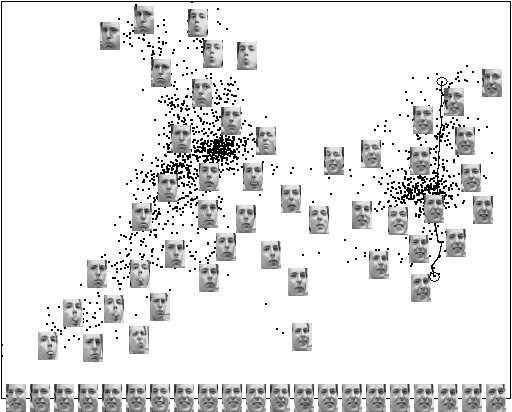}

{\RefSize\RefCLInfoCiteA[Fig.~3]{}{./Images/HeNiyogi-2004-NIPS-LPP}}

\caption{\label{fig:LPPFaces} Dimensionality reduction of face images of a single person down to two dimensions with linear LPP.
  Face images in the plot indicate what some points stand for and the line of faces at the bottom corresponds to the line
  of data points on the right.}

\end{figure}

\markText{An application of LPP to face images of a single person is shown in
  Figure~\ref{fig:LPPFaces}} \cite[]{HeNiyogi-2004}.  Even though the mapping is only
linear, LPP still captures some prominent variations and orders the images nicely in 2D.
The person looks to the left (or right) at the top (or bottom) of the plot, and it smiles
on the right side while it makes faces on the left.

\subsubsection{Nonlinear LPP}

\markText{LPP can be generalized to nonlinear functions by adding a nonlinear
  expansion}\ignoreForEBISS{ \cite[]{Wisk-LN-NonlinearExpansion}} \markText{prior to the
  algorithm.}  Assume $\funcVec(\inVec)$ is such a nonlinear expansion from
$\mathbb{R}^\inDim \to \mathbb{R}^\funcDim$ with $\inDim\ll \funcDim$, then one can
define
\begin{alignat}{3}
  & & & & \eEleLD_{\idxu,\idxx}&\,=\, \funcVec(\inVec_\idxx)^T \funcWVec_\idxu \label{eq:uOff} \\
  &\Longleftrightarrow\quad & & & \eVecLD_\idxu&\,=\, \funcMat^T \funcWVec_\idxu \label{eq:UOfF} \\
  &\text{with}\quad & & & \funcMat&\,:=\, (\funcVec(\inVec_1),\funcVec(\inVec_2),...,\funcVec(\inVec_\Idxx))
\end{alignat}
and then run the algorithm as before.  Notice that now $\funcWVec_\idxu \in\mathbb{R}^\funcDim$
rather than $\mathbb{R}^\inDim$.

Further reading: \cite[]{HeNiyogi-2004}.

\subsection{Spectral clustering}\label{sec:SpectralClustering}

\subsubsection{Objective}

\markText{\emph{Spectral clustering} is an umbrella term for a number of algorithms that
  use the eigenvectors of the Laplacian matrix to perform clustering on a given set of
  data points.} In particular, spectral clustering is often used in image processing to
identify connected parts of a given image and, ideally, identify the extent of the
individual components of an image, a process called \emph{image segmentation}.

As illustrated intuitively in Figure~\ref{Fig:InANutshell-SC} \markText{the eigenvectors
  of the Laplacian matrix place the nodes of connected subgraphs at the same location},
even in two, three, or higher dimensions, if the graph has several subgraphs.  This also
holds for the eigenvectors of the generalized eigenvalue problem, and this also holds
approximately if the subgraphs are not completely separate from each other.
\markText{Given this representation it is much easier than on the original data to
  cluster the nodes with some standard clustering algorithm}\ignoreForEBISS{
  \cite[]{Wisk-LN-Clustering}}.

Remember that for $\numSubs$ intrinsically connected but mutually disconnected subgraphs,
i.e.\ clusters, there are exactly $\numSubs$ eigenvectors with constant values on each of
the clusters.  For extracting $\numSubs$ clusters one would therefore use the first $\numSubs$
eigenvectors, this time including also the first one, see
Property~\myProperty{\ref{item:indicatorVectors}}.

\subsubsection{Algorithm}

\begin{enumerate}
\item[] \markByColor{\textbf{Normalized spectral clustering algorithm}} \cite[]{NgEtAl-2002}
\item Given a set of $\Idxx$ data samples\ignore{ $\{\vec{x}_1, ... \vec{x}_\Idxx\}$},
  \markText{construct a similarity graph $G$} according to one of the methods described
  in Section \ref{section:simGraphs}.  For instance, when performing segmentation on a
  single image, each pixel\ignore{ $\vec{x}_\idxx$} becomes a node of the graph with
  similarity between nodes usually being a function of color and spatial distance.
\item \markText{Compute the} weight matrix $\mat{W}$, degree matrix $\mat{D}$
  (\ref{eq:D_elem_def}), and \markText{Laplacian matrix $\mat{L}$}
  (\ref{EQ_def_laplacian}) \markText{for $G$.}
\item \markText{Compute the first $\numSubs$ eigenvectors of the generalized eigenvalue problem}
  \markByColor{
    \begin{equation}
      \markEqnLect \mat{L} \eVecLD_\idxu = \eValLD_\idxu \mat{D} \eVecLD_\idxu
    \end{equation}
  }\markText{ordered by increasing eigenvalue.}
\item \markText{Arrange the eigenvectors $\eVecLD_1, .., \eVecLD_\numSubs$ in the
    rows\footnotemark\footnotetext{In the original formulation \cite[]{NgEtAl-2002}, the
      vectors were arranged in columns.  We use rows here for consistency with the LEM
      algorithm, see Sec.~\ref{sec:LEM-Algorithm}.}  of a matrix $\mat{U}$ and normalize
    its columns to one to get matrix $\mat{T}$} with
  \begin{equation}
  T_{\idxx\iidxx} = U_{\idxx\iidxx} / \left( \sum_{\idxx'} U^2_{\idxx'\iidxx} \right)^{1/2}
  \end{equation}
  \markText{A $\numSubs$-dimensional representation $\vec{y}_\idxx$ of data sample
    $\idxx$\ignore{$\vec{x}_\idxx$} is now given by the $\idxx$-th column vector of
    $\mat{T}$.}
\item \markText{Perform the $k$-means algorithm} on the set of embedded data points
  $\{\vec{y}_1, ... \vec{y}_\Idxx\}$ to partition the data into $\numSubs$ clusters.
\end{enumerate}

\subsubsection{Sample application}

\markText{Figure~\ref{fig:SCIris} shows an example of applying spectral clustering to an
  old data set} collected by Edgar Anderson \cite[]{Wikipedia-2017-SpectralClustering}.
He measured length and width of the sepal\GWord{Kelchblatt} and petal\GWord{Kronblatt}
from 50 exemplars \markText{of three types of iris.}  One species (red in the left plot)
is well separated from the other two, which in turn are hard to distinguish in the 2D
plots.  Spectral clustering performs fairly well on this task in 4D as one can see by
comparing ground truth on the left with the clustering result on the right.

\begin{figure}[htbp!]
  \centering
  \begin{minipage}[b]{0.5\textwidth}
    \centering
    \includegraphics[width=\textwidth]{./Images/Iris_dataset_scatterplot}
    
    {\RefSize\RefCLInfoCiteA{}{./Images/Iris_dataset_scatterplot}}
  \end{minipage}%
  \begin{minipage}[b]{0.48\textwidth}
    \centering
    \includegraphics[width=\textwidth]{./Images/Specclus_iriscluster}
    
    {\RefSize\RefCLInfoCiteA{}{./Images/Specclus_iriscluster}}
  \end{minipage}
  
  \caption{\label{fig:SCIris} Spectral clustering on iris (the plant, not the eye) data.
    Left: length and width of the sepal\GWord{Kelchblatt} and petal\GWord{Kronblatt} from
    50 exemplars of three types of iris as indicated by the three colors.  Right: Result
    of spectral clustering on the 150 four-dimensional data points.  }
\end{figure}

Further reading: \cite[]{VonLuxburg-2007}, an excellent tutorial on spectral clustering.
\ignore{\cite[]{NaEtAl-2010} and \cite[]{QiEtAl-2016} for current articles on (efficient)
$k$-means clustering.}

\paragraph{Acknowledgments:} We thank Jan Melchior and Merlin Sch\"uler for valuable
feedback on an earlier version of these lecture notes.










\bibliography{references,/home/wiskott/Lehre/Modules/lectureNotes}